\documentclass[10pt]{article} 
\usepackage[preprint]{tmlr}

\usepackage{algorithm}
\usepackage{algpseudocode}

\usepackage{hyperref}
\usepackage{array}
\usepackage{fullpage}
\usepackage{adjustbox}
\usepackage{lineno}

\usepackage{url}
\usepackage{xcolor}

\usepackage{multirow}
\usepackage{threeparttable}
\usepackage{makecell}
\usepackage{booktabs} 

\usepackage{amsmath,amsfonts,bm}

\def\eqref#1{Equation~(\ref{#1})}

\def\1{\bm{1}}

\RequirePackage{amsmath}
\RequirePackage{amssymb}
\RequirePackage{amsthm}
\RequirePackage{bm} 
\RequirePackage{url}
\usepackage{multirow}
\usepackage{natbib}
\usepackage{graphicx}
\usepackage{subfigure}
\usepackage{makecell}
\usepackage{booktabs}
\usepackage{array}
\usepackage{url}

\usepackage{dsfont}

\usepackage{enumerate}
\usepackage[OT1]{fontenc}
\usepackage{natbib}

\usepackage{mathrsfs}

\usepackage{xcolor}
\usepackage{hyperref}
\usepackage{float}  
\usepackage{colortbl}

\def\##1\#{\begin{align}#1\end{align}}
\def\$#1\${\begin{align*}#1\end{align*}}

\newcommand{\E}{\mathbb{E}}

\usepackage{xcolor}

\definecolor{red1}{HTML}{f47983}
\definecolor{blue1}{HTML}{3eede7}
\definecolor{yellow1}{HTML}{f5dd6f}

\newtheorem{example}{Example}
\newtheorem{remark}{Remark}

\title{Reinforce-Ada: An Adaptive Sampling Framework under Non-linear RL Objectives}

\author{Wei Xiong\thanks{Equal contribution. A detailed attribution of authorship credits is provided in Appendix~\ref{appendix:credits}. 
 Correspondence to Hanze Dong (\url{hanzedong@microsoft.com})
and Wei Xiong (\url{wx13@illinois.edu}).}$\ ^\circ$ \quad Chenlu Ye$^{*\circ}$ \quad Baohao Liao$^{*\blacksquare}$ \quad Hanze Dong$^{*\diamondsuit}$ \\
Xinxing Xu$^{\diamondsuit}$\quad
Christof Monz$^{\blacksquare}$\quad
Jiang Bian$^{\diamondsuit}$\quad
Nan Jiang$^{\circ}$\quad
Tong Zhang$^{\circ}$
\\\\
\textnormal{\emph{$^\circ$University of Illinois Urbana-Champaign
\quad$^\diamondsuit$Microsoft Research\quad$^\blacksquare$University of Amsterdam
}}}

\def\month{MM}  
\def\year{YYYY} 
\def\openreview{\url{https://openreview.net/forum?id=XXXX}} 

\begin{document}

\maketitle

\begin{abstract}
Reinforcement learning (RL) for large language model reasoning is frequently hindered by \textit{signal loss}, a phenomenon where standard uniform sampling with small group sizes fails to uncover informative learning signals for difficult prompts. We demonstrate that this collapse is  a statistical artifact of undersampling rather than an inherent model limitation. To address this systematically, we introduce a theoretical framework based on optimizing a non-linear RL objective (e.g., log-likelihood). We show that this objective naturally induces a weighted gradient estimator that prioritizes difficult prompts, which can be robustly realized through adaptive sampling. Guided by this framework, we propose \textsc{Reinforce-Ada}, a family of algorithms that dynamically allocates inference budgets based on prompt difficulty, effectively scaling up RL compute to where it is needed most. Unlike passive filtering methods that discard low-signal prompts, \textsc{Reinforce-Ada} actively invests compute to recover them. We introduce two efficient realizations: an estimation-based approach and a model-free sequential sampling approach. Extensive experiments across multiple benchmarks show that \textsc{Reinforce-Ada} significantly outperforms uniform baselines like GRPO, recovering lost signals and accelerating convergence by up to $2\times$ while maintaining the same total inference budget. Code is available at \url{https://github.com/RLHFlow/Reinforce-Ada}.
\end{abstract}

\begin{figure}[h!]
    \centering
    \includegraphics[width=0.45\linewidth]{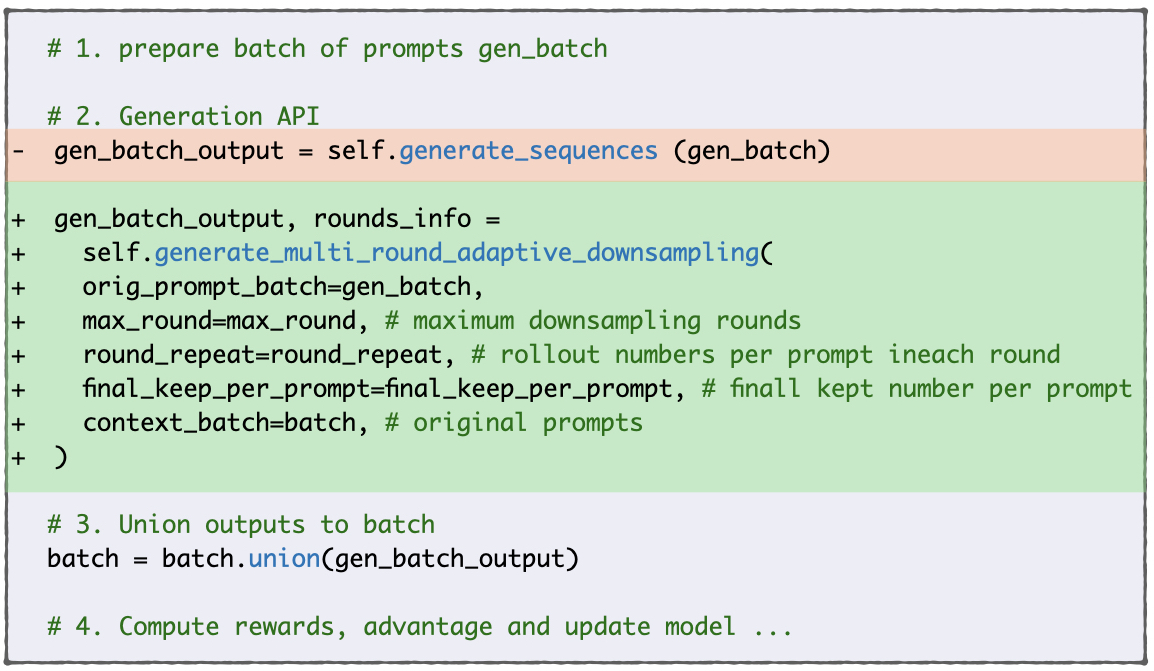}
        \includegraphics[width=0.54\linewidth]{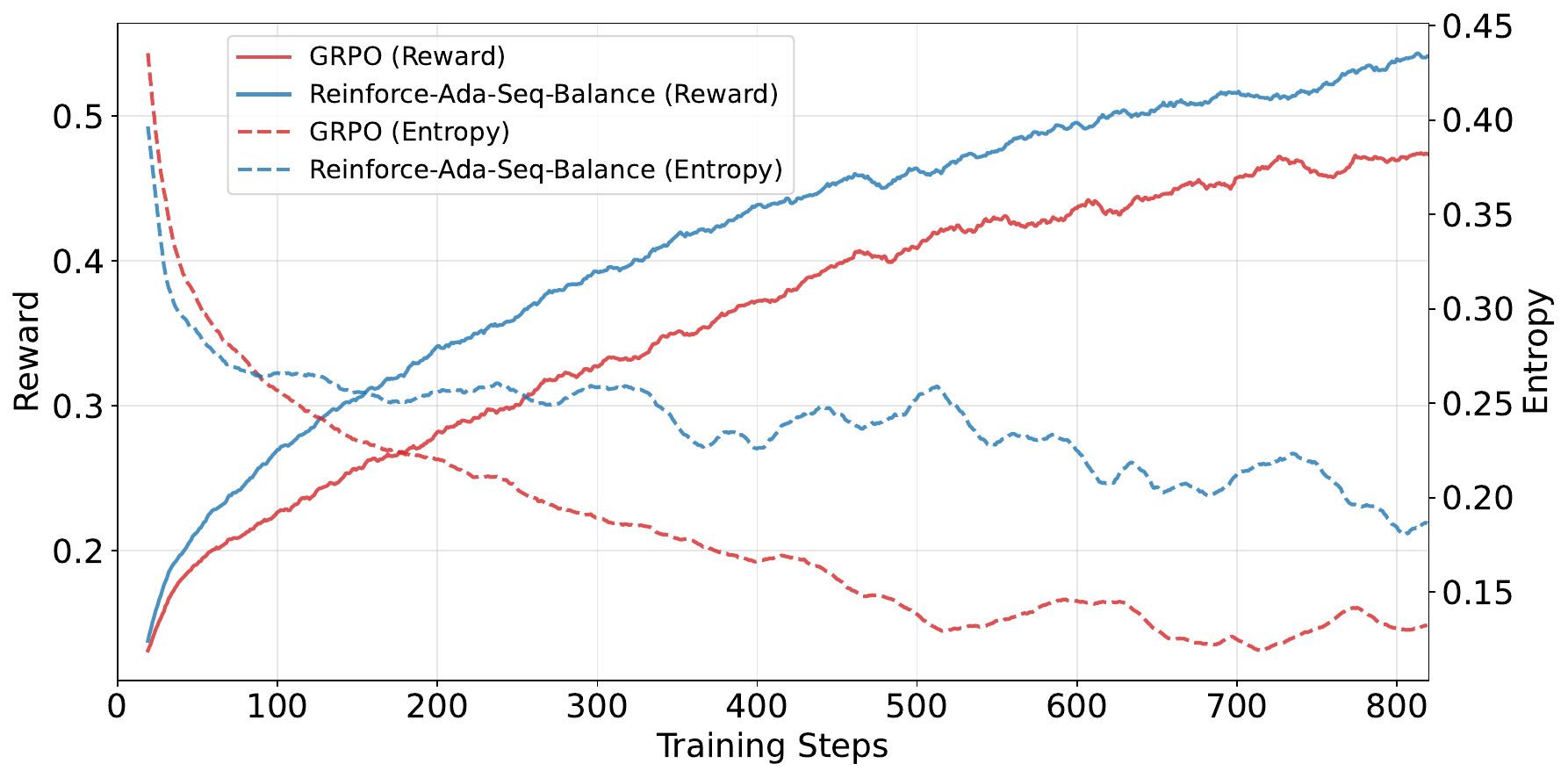}
    \caption{Plug-and-play usage. Left: a direct replacement of the generation API in \texttt{verl} (\texttt{generate\_sequences} $\rightarrow$ \texttt{generate\_multi\_round\_adaptive\_downsampling}). Right: with no other changes, \textsc{Reinforce-Ada} attains faster reward growth and a higher asymptote than GRPO.}
    \label{fig:code}
\end{figure}

\section{Introduction}

Reinforcement learning (RL) has become a central paradigm for aligning large language models (LLMs). 
Conceptually, the training objective is to maximise the expected reward of model outputs under a given prompt distribution. The primary challenge is not from the formulation of the objective but from the high variance in gradient estimates, which introduces instability into the optimization process. 

Formally, for a prompt $x \sim d_0$, the policy $\pi_\theta$ produces a response $a \sim \pi_\theta(\cdot|x)$, and a verifier yields a binary reward $r^\star(x,a)\in \{0,1\}$. The learning objective is
\begin{equation}\label{eqn:target}
J(\theta) = \mathbb{E}_{x \sim d_0} [\E_{a \sim \pi_\theta(\cdot|x)} r^\star(x, a)] =: \mathbb{E}_{x \sim d_0} p_\theta(x),
\end{equation}
where $p_\theta(x)$ is often referred to as the pass rate, which is metric of the prompt difficulty given the model $\pi_\theta$.

Estimating its gradient requires sampling multiple responses. With only a few samples per prompt, inference is affordable but the gradient is noisy; with many samples, the signal is clear but inference becomes prohibitively expensive. The trade-off between signal quality and cost is a central challenge in RL for LLMs.

The vanilla policy gradient with small $n$ has notoriously high variance. A standard remedy introduces a reward baseline $b(x)$, yielding $$g_\theta(x,a) = (r^\star(x,a) - b(x)) \cdot \nabla_\theta \log \pi_\theta(a|x),$$ which stabilizes training while preserving unbiasedness.

Group Relative Policy Optimization (GRPO) \citep{shao2024deepseekmath} extends this principle by assigning $n$ responses per prompt and normalizing each sample’s advantage:
\begin{equation}\label{eqn:adv_grpo}
A_{\text{GRPO}}(x,a_i) = \frac{r_i - \bar{r}}{\sigma_r + \varepsilon},
\end{equation}
where $\bar{r}$ and $\sigma_r$ are the mean and standard deviation of group rewards. This group-wise normalization highlights informative variations while suppressing noise, making GRPO widely adopted in practice.

Despite these benefits, GRPO fundamentally relies on a small and fixed $n$ per prompt, creating a vulnerability to signal collapse. When all $n$ samples for a prompt yield identical rewards, either all correct or all incorrect, the group mean $\bar{r}$ equals each reward $r_i$.
Consequently, the computed advantages vanish ($A_{\text{GRPO}}=0$), resulting in a zero gradient.
Such uniform-reward scenarios arise frequently: during early training phases when the model fails on all attempts for challenging prompts, and later when it consistently succeeds on trivial ones. For instance, with a group size of 8, if the pass rate $p=0.1$, the probability that all 8 samples are 0 is $0.9^8 \approx 19\%$. Empirically, even with $n=32$, more than half of prompts fall into this ``zero-gradient'' regime as models improve \citep{yu2025dapo}.

\begin{figure}[h]
     \centering
\includegraphics[width=0.45\textwidth]{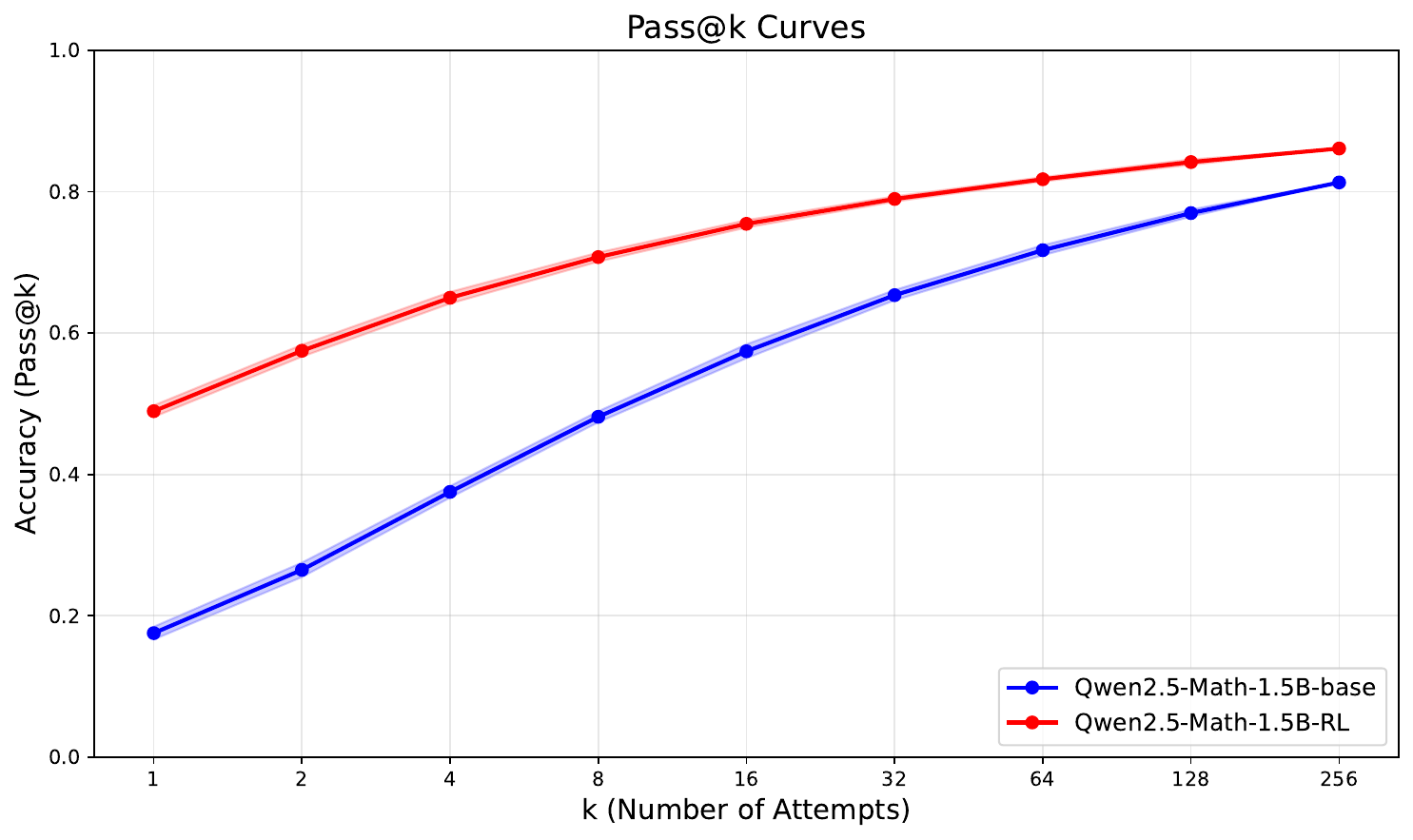}
\includegraphics[width=0.45\textwidth]{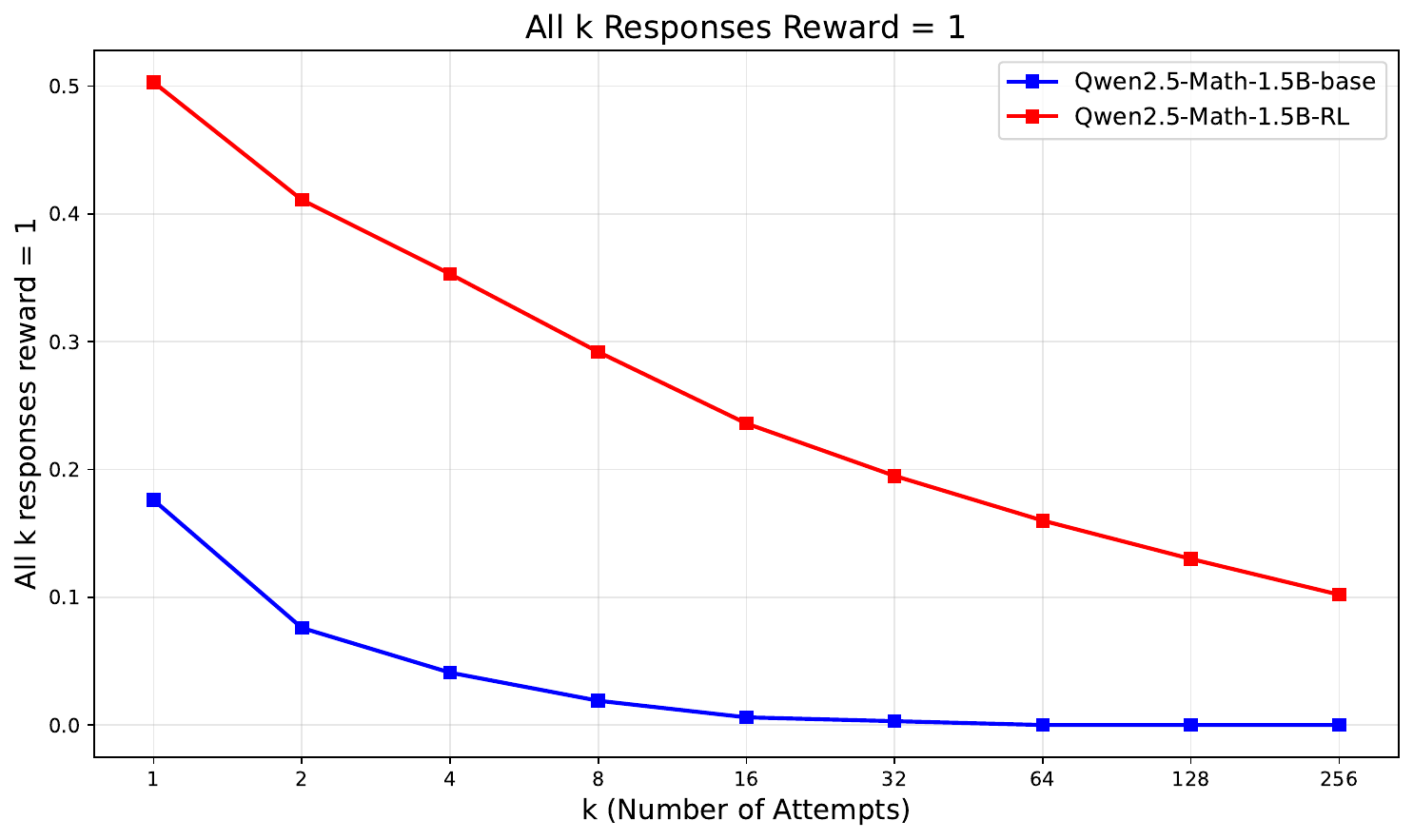}
     \caption{
Pass@k curves (left) and the ratio of prompts with all-correct responses (right) for two models on a subset of the Open-R1 prompt set. 
The models tested are the Qwen2.5-Math-1.5B base model and an intermediate checkpoint from its RL training. The percentage of prompts yielding all-correct/all-incorrect responses is high for small $k$ but drops significantly as $k$ increases. This suggests that signal loss is often a statistical artifact of small sample groups.
}    \label{fig:pass_at_n}
 \end{figure}

Crucially, this collapse is not due to the prompts being inherently trivial or impossible, but a \textbf{statistical artifact} of undersampling ($\hat{p}_\theta(x)=0 \text{ or } 1$: estimating $p_\theta(x)$ with finite samples from $\pi_\theta(\cdot|x)$). Training prompts are typically filtered to ensure moderate difficulty (e.g., \citet{yang2024qwen2} retain prompts where 2–5 out of 8 responses are correct), meaning most prompts have non-trivial success probabilities ($0<\hat{p}_\theta<1$). With small $n$, however, random fluctuations make it likely to observe all-correct or all-incorrect groups, thereby masking the true learning signal. Larger $n$ reliably recovers these signals, but at unsustainable inference cost (Figure~\ref{fig:pass_at_n}\footnote{This figure shows results on Open-R1 subset for Qwen2.5-Math-1.5B and an RL-trained checkpoint (step 400). The base model scores 26.5\% at pass@1, but its pass@256 rises to 81.3\%, highlighting its potential to solve the majority of prompts. Similarly, the model exhibits 35.3\% all-correct groups at $n=4$, but only 10.2\% at $n=256$. These results demonstrate that the missing signal is often recoverable with larger $n$, confirming that uniform-reward collapse is a statistical artifact of undersampling rather than a model limitation.}).  

To resolve this trade-off, we first introduce a general framework that provides a principled foundation for our solution. We argue that the standard RL objective in \eqref{eqn:target} is the root of the problem, as it values all prompts equally regardless of their difficulty. However, from a learning dynamics perspective, solving a difficult prompt (where the model currently fails) provides significantly more information than repeatedly solving an easy one. To effectively balance cost and signal discovery, we must prioritize prompts based on their pass rates. This motivates us to consider a non-linear objective $J_f(\theta) = \mathbb{E}_x [f(p_\theta(x))]$, such as the log-likelihood $f(p) = \log p$. The gradient of this objective naturally acquires a prompt-dependent weight, $\nabla J_f = \E_x [f'(p) \cdot \nabla p]$. This weight (e.g., $1/p$ for the log objective) explicitly assigns more importance to difficult prompts, directly targeting the signal loss problem. This formulation unifies two implementation paths: explicitly weighting the gradient, or implicitly re-weighting by allocating more samples $n_i$ (i.e., adaptive sampling).

Building on this framework, we propose \textsc{Reinforce-Ada}, a family of adaptive sampling algorithms that robustly implement this implicit weighting principle. We present two efficient realizations to determine the optimal budget allocation:
\begin{enumerate}
\item \textbf{Estimation-based Allocation (\textsc{Reinforce-Ada-Est})}: This variant estimates pass rates online using a lightweight value network or Bayesian moving averages, and then explicitly allocates sampling budgets proportional to the theoretical target (e.g., $n_i \propto 1/\hat{p}_i$).
\item \textbf{Implicit Allocation via Sequential Sampling (\textsc{Reinforce-Ada-Seq})}: This variant adopts a model-free approach inspired by multi-armed bandits. It continues sampling until sufficient signal is found (e.g., until $K$ positive responses are collected). We show that this simple stopping rule naturally achieves the desired allocation without needing explicit estimation.
\end{enumerate}

\textsc{Reinforce-Ada} is a plug-and-play replacement for the generation step in standard RL pipelines, requiring no architectural modifications. Unlike prompt selection or curriculum methods that operate at the macro prompt level, our method performs micro-level response allocation, shaping the internal structure of each training group. Across multiple LLMs and benchmarks, it consistently improves signal quality and sample efficiency, achieving the benefits of large-$n$ training at a fraction of its cost.

\section{The Weighted Objective Framework}
\label{sec:theory}

\subsection{Prior Approach: Passive Filtering and Large Group Size}

The standard RL objective is $J(\theta) = \E_{x \sim d}[p_\theta(x)]$. In practice, $d$ is often a simple uniform distribution over the training set. The most direct, unbiased estimator for this objective is to sample a batch of prompts and a uniform group size $n$ of responses for each.

However, this naive, uniform-sampling approach, leads to \textbf{signal loss} when $n$ is small: if all responses in a group yield the same reward (all correct or all incorrect), the advantages normalize to zero, and the gradients vanish (see \eqref{eqn:adv_grpo}). Crucially, this ``signal collapse'' is not due to the prompts being inherently trivial or impossible, but is a statistical artifact of undersampling.

Prior work has observed this issue and proposed \textit{passively filtering-out} groups with uniform rewards \citep{yu2025dapo, xiong2025minimalist}. 
While this prevents wasted gradient computations, it still incurs the significant upfront cost of generating responses that are ultimately discarded. Moreover, if difficult prompts are systematically discarded due to a lack of positive signal, the model never learns to solve them, thus hindering training improvement.

With these observations, a natural alternative is to use a large $n$ to reliably capture learning signals. Indeed, concurrent work by \citet{hu2025brorlscalingreinforcementlearning} validates this intuition, showing that increasing $n$ up to 512 recovers valid signals and improves performance. However, generating hundreds of samples for every prompt is computationally prohibitive at scale. As demonstrated by DeepSeek-R1 \citep{deepseekai2025deepseekr1incentivizingreasoningcapability}, we can use only $n=16$ responses per prompt to get an effective gradient for model updates. This reveals a significant gap between the inference budget needed to reliably find a learning signal and the update budget required for an effective parameter update.

To bridge this gap, we need a framework that moves beyond uniform allocation. We propose an adaptive approach that smartly allocates a larger inference budget to prompts where the signal is scarce, efficiently discovering robust learning signals without the waste of the uniform, large-$n$ approach.

\subsection{A Principled Solution: The Weighted Objective Framework} \label{sec:principle}
Beyond practical intuition, we establish a principled theoretical foundation for allocating inference budgets adaptively. We argue that the root cause of the uniform sampling inefficiency is the standard RL objective $J(\theta) = \mathbb{E}_x[p_\theta(x)]$, which values all prompts equally.

Instead, we propose a general theoretical framework based on optimizing \textbf{a non-linear transformation} of the pass rate:
\begin{equation} \label{eqn:general}
    J_f(\theta) = \E_{x \sim d_0} f(p_\theta(x)),
\end{equation}
where $f:  \to \mathbb{R}$ is a non-decreasing function. When $f(p) = p$, we recover the standard objective. We also may want to choose a concave $f$. In this way, we implicitly assign higher marginal utility to improvements on prompts with low pass rates (difficult prompts). The policy gradient for this general objective reveals a crucial insight. By applying the chain rule, the gradient acquires a prompt-dependent weight:
$$\nabla J_f(\theta) = \mathbb{E}_{x \sim d_0} \Big[\underbrace{f'(p_\theta(x))}_{\text{Weight } w(p)} \cdot \mathbb{E}_{a \sim \pi_\theta(\cdot|x)} \Big[ \nabla_\theta \log \pi_\theta(a|x) \cdot r(x, a) \Big]\Big].$$
A canonical choice that targets signal loss is the \textbf{log-likelihood objective}, $f(p) = \log p$. This yields a weight $w(x) = 1/p_\theta(x)$, which scales inversely with the model's current performance. This explicitly instructs the optimizer to prioritize difficult prompts (where $p \to 0$) with a potentially large weight, counteracting the vanishing gradients typically observed in these regimes.
\begin{example}[Log objective $f(t)=\log(t)$] Here $J_{\log}(\theta) = \E_{x \sim d_0} \log(p_\theta(x))$, and the gradient for a prompt $x$ is
$$
\frac{1}{p_\theta(x)} \cdot \E_{a \sim \pi_\theta(\cdot|x)} \Big[ \big( r^\star(x, a)  \big)\cdot \nabla_\theta \log \pi_\theta(a|x)\Big].
$$
\end{example}
This explicitly instructs the optimizer to prioritize difficult prompts (where $p \to 0$) with a potentially infinite weight. We may also consider other functions such as the power function to present a softer weight.

\begin{example}[Power function $f(t)=t^\alpha$, $\alpha>0$] In this case, $J_{\alpha}(\theta) = \E_{x \sim d_0} [p_\theta(x)]^\alpha$, and the gradient for prompt $x$ is
$$
\alpha p_\theta(x)^{\alpha-1} \cdot \E_{a \sim \pi_\theta(\cdot|x)} \Big[ \big( r^\star(x, a)  \big)\cdot \nabla_\theta \log \pi_\theta(a|x)\Big].
$$
Setting $\alpha=\frac{1}{2}$ gives, 
$$
\frac{1}{2\sqrt{p_\theta(x)}} \cdot \E_{a \sim \pi_\theta(\cdot|x)} \Big[ \big( r^\star(x, a)  \big)\cdot \nabla_\theta \log \pi_\theta(a|x)\Big].
$$
\end{example}

\begin{figure}[htp]
    \centering
    \includegraphics[width=0.45\textwidth]{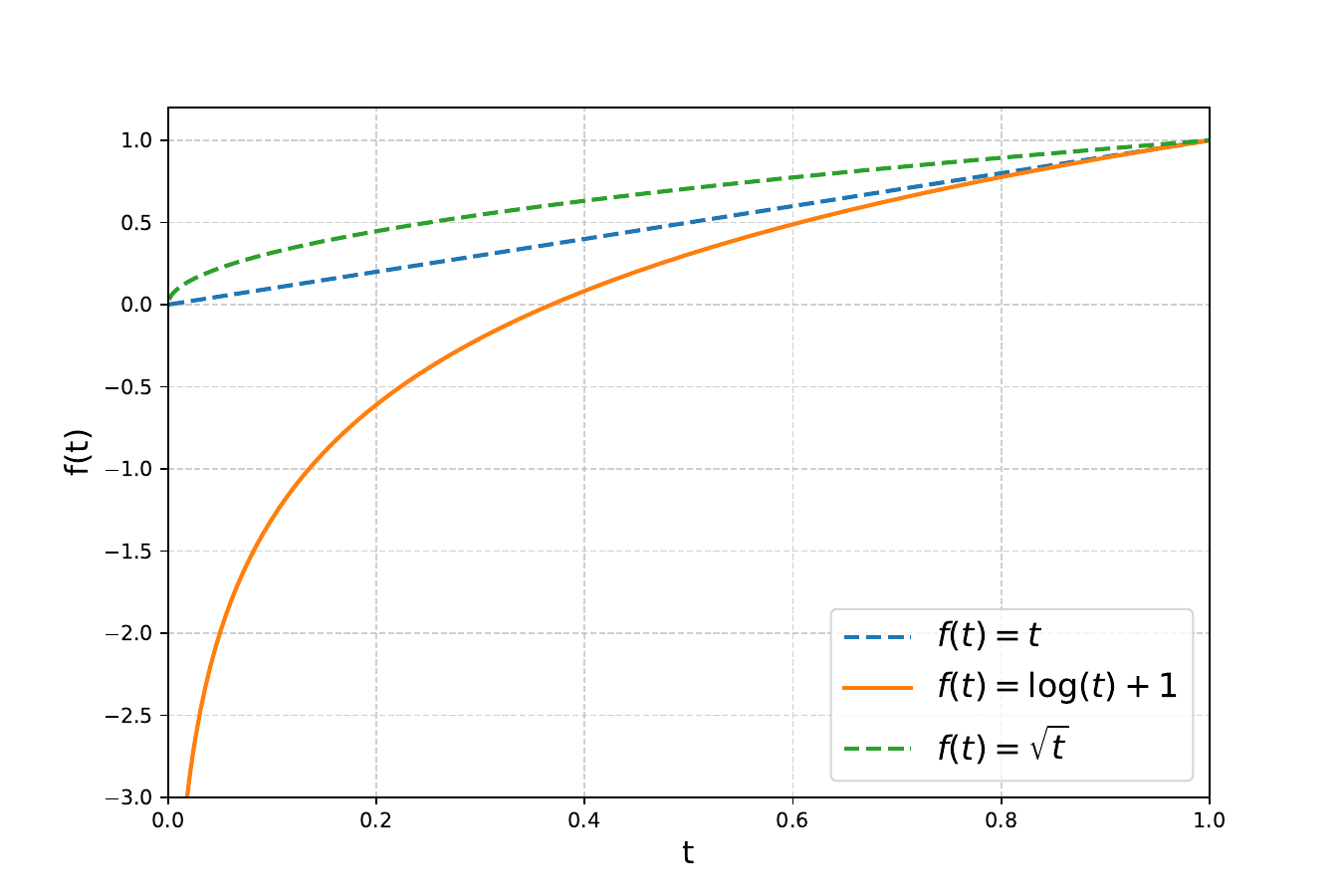}  
\includegraphics[width=0.45\textwidth]{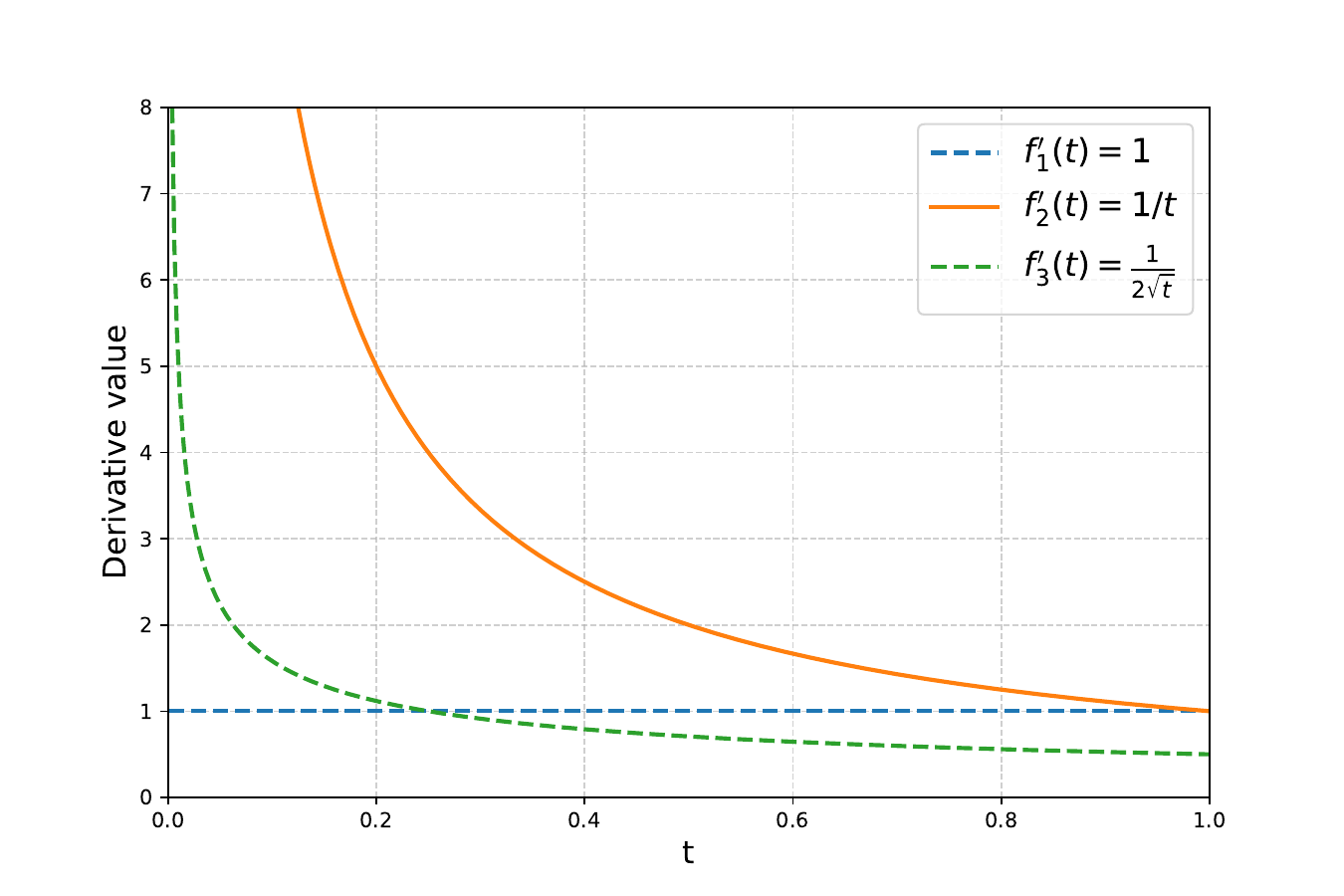} 
\caption{Visualization of different $f(t)$ and $f'(t)$. The concave functions $\log (t)$ and $\sqrt{t}$ assign larger weights $f'(t)$ to difficult prompts ($t\rightarrow0$).
}
        \label{fig:objective} 
\end{figure}

This weight $w(p) = f'(p)$ formalizes our intuition. We present a visualization of these function choices in Figure~\ref{fig:objective}. For the standard objective $f(p)=p$, the weight is $w(p)=1$, , confirming that it treats all prompts uniformly. In contrast, both the logarithmic and power objectives introduce prompt-dependent weights $w(p)$ provide us with a weighted gradient, naturally leading to an adaptive treatment. In particular, for concave functions such as $\log t$ and $\sqrt{t}$ (on $(0, 1)$), we observe that their prompt-dependent weights decrease with $t$, assigning larger weights to more difficult prompts. We will also provide an interesting explanation of GRPO's weight under the log-objective in Appendix~\ref{appendix:vr}.

\paragraph{Implementation of Weighted Objective: Explicit vs. Implicit.} This weighted gradient formulation $\mathbb{E}_x [w(p) \cdot \nabla p(x)]$ unifies distinct implementation strategies. We can decompose the total weight into two components, $w(p) = w_{\mathrm{sample}}(p) \cdot w_{\mathrm{grad}}(p)$, allowing for a flexible spectrum of solutions:
\begin{enumerate}
\item \textbf{Explicit Weighting ($w_{\mathrm{sample}}=1, w_{\mathrm{grad}}=w$):} One retains uniform sampling ($n_i = n$) but explicitly multiplies the gradient by $w(\hat{p}_i)$. However, this strategy fails to solve the core ``signal loss'' problem. Since we rely on Monte-Carlo estimation for the pass rate, a small group size $n$ will often yield $\hat{p}_i = 0$ for difficult prompts, causing the gradient to vanish rather than be weighted up.
    \item \textbf{Implicit Weighting via Sampling ($w_{\mathrm{sample}}=w, w_{\mathrm{grad}}=1$):} Alternatively, we can absorb the weight $w(p)$ into the sampling distribution. We sample prompts from a reweighted distribution $d'(x) \propto d_0(x) \cdot w(p(x))$ and apply the standard, unweighted gradient. In a batched setting with group sampling, this is naturally implemented by assigning a \textbf{variable group size} $n_i \propto w(p_i)$ to each prompt.
    \item \textbf{Hybrid Strategy:} We can balance the two approaches. For instance, with the log-objective ($w(p)=1/p$), a pure sampling strategy ($n_i \propto 1/p$) might be too aggressive, allocating excessive compute to the hardest prompts. A balanced approach is to allocate samples proportional to $1/\sqrt{p}$ and apply a residual explicit weight of $1/\sqrt{p}$ to the gradient. This hybrid approach achieves a ``sweet spot'' by dividing the variance into two stages.
\end{enumerate}
This framework highlights that the intuitive adaptive group sizes are not just a heuristic, but a robust implementation of a non-linear objective. By allocating $n_i \propto 1/p_i$ (Implicit Weighting), we naturally direct more computational budget to difficult prompts. This explicitly solves the signal loss problem: the increased budget allows us to recover rare signals that a uniform small-$n$ approach would miss.

\paragraph{Theoretical Connections with GRPO and Variance-reduction Gradient Estimation.} This framework is broadly applicable and offers a unified perspective on recent advancements. As a side product, it explains why algorithms like GRPO \citep{shao2024deepseekmath} utilize prompt-dependent weights in their advantage and gradient formulations. Furthermore, we find that the log-objective is essential for deriving the optimal variance-reduction gradient estimator under a fixed inference budget. In Appendix~\ref{appendix:vr}, we generalize the findings of \citet{yao2025optimizing} from rejection sampling fine-tuning to the general Reinforce algorithm, proving the optimality of the log-objective for efficient training.

\section{Reinforce-Ada: Reinforce with Adaptive Sampling}
\label{sec:algorithm}
\subsection{Reinforce-Ada-Est: Estimation-based Allocation}

Our theoretical framework suggests allocating an implicit sampling budget $n_i \propto 1/p_i$ to optimize the log-objective. The central challenge, however, is that the true pass rate $p_i$ is unknown prior to sampling.

A common approach in related work \citep{yao2025optimizing} follows an ``explore-then-exploit'' paradigm. In these methods, pass rates ($\hat{p}_i$) are first estimated with a small ``explore'' budget, and then a larger online sampling budget is allocated proportionally. This two-stage process, however, suffers from a critical flaw. The initial estimation, based on a limited budget, has high variance. This error is most severe for difficult prompts (low $p_i$), where a small budget will almost certainly yield zero positive responses. This leads to a catastrophic estimate of $\hat{p}_i = 0$, causing the algorithm to discard the very prompts that need the most attention.

Consequently, any allocation rule (like $1/\sqrt{\hat{p}_i}$) will either fail (divide by zero) or assign zero budget, forcing the algorithm to discard these prompts. This means the two-stage approach ends up replicating the same problem as passive filtering (like DAPO \citep{yu2025dapo}), failing to solve the very signal loss problem it was designed for. 

\begin{algorithm}[t]
\caption{Reinforce-Ada-Est under Log-objective (One Training Iteration)}
\label{alg:ada_est}
\begin{algorithmic}[1]
\State \textbf{Input:} Current policy $\pi_\theta$, batch of prompts $\mathcal{D}$, effective update size $n$, sampling bounds $[N_{\min}, N_{\max}]$, total budget target $N_{\text{total}}$, and estimator $\mathcal{E}$

\State \textcolor{blue}{\textsc{Phase 1: Estimation and Budget Allocation}}
\State Initialize response set $\mathcal{S}_x \leftarrow \emptyset$ for all $x \in \mathcal{D}$
\State Get current pass rate estimates $\hat{p}_x \leftarrow \mathcal{E}(x)$ for all $x \in \mathcal{D}$
\State \textbf{Allocate budgets:} Set $N_x \propto 1/\sqrt{\hat{p}_x + \epsilon}$ subject to $N_x \in [N_{\min}, N_{\max}]$, targeting total usage $\sum N_x \approx N_{\text{total}}$
\For{each prompt $x \in \mathcal{D}$}
    \State Sample $N_x$ responses $\{a_j, r_j\}_{j=1}^{N_x} \sim \pi_\theta(\cdot | x), r^\star(\cdot, \cdot)$
    \State Store collection: $\mathcal{S}_x \leftarrow \{a_j, r_j\}_{j=1}^{N_x}$
\EndFor

\State {\textcolor{blue}{\textsc{Phase 2: Training Batch and Objective Construction}}}
\State Initialize an empty set for the final training data: $\mathcal{B} \leftarrow \emptyset$
\For{each prompt $x \in \mathcal{D}$}\Comment{\textcolor{blue}{Use global statistics for baseline and weight}} 
    \State Let $\mathcal{S}_x$ be the full set of collected samples for prompt $x$
    \State Compute global mean (high-fidelity pass rate): $\bar{r}_x \leftarrow \frac{1}{|\mathcal{S}_x|}\sum_{j=1}^{|\mathcal{S}_x|} r_j$      
    \State Compute explicit weight: $\alpha_x \leftarrow 1/\sqrt{\bar{r}_x + \epsilon}$ \Comment{\textcolor{blue}{Apply residual weight for log-objective}}
    \State Compute weighted advantage for $i$-th response: $A_i \leftarrow (r_i - \bar{r}_x) \cdot \alpha_x$
\EndFor
\State Compute the policy gradient objective with batch $\mathcal{B}$ and update the model.
\State Update estimator $\mathcal{E}$ with newly collected statistics $\{\mathcal{S}_x\}_{x \in \mathcal{D}}$.
\end{algorithmic}
\end{algorithm}

\paragraph{Adaptive Sampling with Pass Rate Estimation.} To overcome this, we propose \textsc{Reinforce-Ada-Est}, which estimates pass rates online using historical training data or auxiliary models, rather than a separate exploration phase. We introduce two strategies for this estimation. A natural idea is to train a value network $V_\phi(x)$ (similar to the critic) to predict the pass rate. Unlike standard PPO which requires estimating token-level values, our task is simpler: we only need the prompt-level expected success rate. Meanwhile, the estimator does not need to be perfectly accurate; it only needs to capture the relative difficulty of prompts to guide budget allocation.

The second strategy is \textsc{Ada-EMA}, a 
model-free Bayesian Moving Average approach, which is inspired by \citet{qu2025can}. Suppose we encounter the prompt across different training epochs $t=1, 2, \dots$. We can therefore maintain a running statistic for each prompt to track its difficulty without an auxiliary model.
Crucially, since the policy $\pi_\theta$ is constantly updating, the pass rate is non-stationary; data collected in early steps becomes stale. To handle this, we apply an exponential decay to historical counts before integrating new data. Specifically, suppose at step $t$, we allocate a budget of $n_t$ samples to prompt $x$ and observe $k_t$ successes. We update the running counters as follows:
$$N_{total}^{(t)} \leftarrow \lambda N_{total}^{(t-1)} + n_t, \quad N_{pos}^{(t)} \leftarrow \lambda N_{pos}^{(t-1)} + k_t,$$
where $\lambda \in (0, 1)$ is a discount factor that weighs recent data more heavily. The pass rate is then estimated using a Bayesian approach with a prior Beta($\alpha$,$\beta$): $\hat{p}_t = (N_{pos}^{(t)} + \alpha) / (N_{total}^{(t)} + \alpha + \beta)$. This ensures that even if a prompt yields zero successes in the current batch ($k_t=0$), the estimate remains non-zero due to the prior and historical accumulation, preventing the algorithm from permanently discarding difficult prompts.

We will mainly implement the \textsc{Reinforce-Ada-Est} via the hybrid strategy under the log objective, as this divide the variance into two separate stages and may be more stable. Specifically, we allocate the inference budget proportional to $1/\sqrt{p_i}$ and leave another $1/\sqrt{p_i}$ to the advantage. We also include an ablation in Appendix~\ref{appendix:abl_hybrid}.

\subsection{Reinforce-Ada-Seq: Implicit Allocation via Sequential Sampling}

While the estimation-based strategies (\textsc{Reinforce-Ada-Est}) provide a direct way to allocate budgets, they introduce specific implementation requirements. For instance, training an auxiliary value network adds computational overhead and architectural complexity to the RL pipeline. Similarly, the Bayesian approach relies on maintaining persistent statistics and revisiting the same prompts multiple times, which may not suit all training setups.

\begin{algorithm}[t]
\caption{Reinforce-Ada-Seq (One Training Iteration)}
\label{alg:grpo_adaptive_simplified}
\begin{algorithmic}[1]
\State \textbf{Input:} Current policy $\pi_\theta$, batch of prompts $\mathcal{D}$, effective group size for update $n$, number of sampling rounds $N$, samples per round $M \geq n$, and exit condition function $\text{ExitCondition}(\cdot)$

\State \textcolor{blue}{\textsc{Phase 1: Adaptive Sampling Data Collection}}
\State Set all prompts $x$ as active and initialize response set $\mathcal{S}_x \leftarrow \emptyset$

\For{$t = 1, \dots, N$} \Comment{\textcolor{blue}{Iterate through sampling rounds}}
    \For{each prompt $x \in \mathcal{D}$ where $\text{active}(x)$ is true}
            \State Sample $M$ responses $\{a_j, r_j\}_{j=1}^M \sim \pi_\theta(\cdot | x), r^\star(\cdot, \cdot)$
        \State Add to collection: $\mathcal{S}_x \leftarrow \mathcal{S}_x \cup \{a_j, r_j\}_{j=1}^M$
        \If{$\text{ExitCondition}(\mathcal{S}_x)$ is met}
            \State   Mark prompt as inactive: $\text{active}(x) \leftarrow \text{false}$
        \EndIf
    \EndFor
\EndFor

\State {\textcolor{blue}{\textsc{Phase 2: Training Batch and Objective Construction}}}
\State Initialize an empty set for the final training data: $\mathcal{B} \leftarrow \emptyset$
\For{each prompt $x \in \mathcal{D}$}\Comment{\textcolor{blue}{Use all collected samples (``global statistics") for normalization}} 
    \State Let $\mathcal{S}_x = \{(a_j, r_j)\}_{j=1}^{|\mathcal{S}_x|}$ be the full set of collected samples for prompt $x$
    \State Compute global mean: $\bar{r}_x \leftarrow \frac{1}{|\mathcal{S}_x|}\sum_{j=1}^{|\mathcal{S}_x|} r_j$     
\State Form update group by downsampling $\mathcal{S}_x$ to size $n$, trying to ensure $\ge n/2$ size for each correct or incorrect subset (fill from the other if needed). \Comment{\textcolor{blue}{Downsample to create the effective group}}
    \State Compute advantage for $i$-th response of prompt $x$ as $A_i \leftarrow r_i - \bar{r}_x$.
\EndFor
\State Compute the policy gradient objective with batch $\mathcal{B}$ and update the model.
\end{algorithmic}
\end{algorithm}

Motivated by these considerations, we design an alternative strategy that achieves adaptive allocation without explicitly estimating the pass rate. We design a model-free algorithm that integrates estimation and sampling into a unified online process. Our core idea leverages a simple statistical property: if we keep sampling responses until we collect a fixed number of correct answers (say, $K=1$), the expected total number of samples $\mathbb{E}[N_i]$ is exactly $1/p_i$. This provides a ``built-in'' mechanism to automatically achieve the $n_i \propto 1/p_i$ allocation required by our log-objective, without needing to estimate $p_i$ beforehand.

We implement a more general version of this ``sample-until-condition'' idea, inspired by \textit{successive elimination methods} in multi-armed bandit literature \citep{slivkins2019introduction}. We present the code of a single training step of this method in Algorithm~\ref{alg:grpo_adaptive_simplified}. Specifically, the algorithm operates over a batch of prompts in sequential rounds:
\begin{enumerate}
    \item \textbf{Initialization:} All prompts in the current batch begin in an \textit{active set} (active arms).
    \item \textbf{Iterative Sampling:} In each round, we generate $M$ new responses (e.g., $M=16$) for every prompt currently in the \textit{active set}.
    \item \textbf{Elimination:} At the end of each round, active prompts are checked against an exit condition, and those that satisfy it are removed from future rounds.  \footnote{We also experimented with a more complex variant that estimates pass rates and allocates budgets proportionally within each round. However, this did not yield clear performance gains, so we use this simpler, easy-to-implement version.}
\end{enumerate}
This sequential sampling process continues until all prompts are resolved or a maximum budget (e.g., $N_{\text{max}}=128$) is reached.

\paragraph{Exit condition.} The elimination rule plays a central role in shaping the algorithm's behavior. We consider two primary exit conditions:
\begin{itemize}
\item \textbf{Positive-focused} (\textsc{Reinforce-Ada-Seq-pos}): A prompt is deactivated once we collect at least $K_{\text{pos}}$ correct responses (e.g., $K_{\text{pos}}=16$).
\item \textbf{Balanced} (\textsc{Reinforce-Ada-Seq-balance}): A prompt is deactivated once at least $K_{\text{pos}}$ correct \textbf{and} $K_{\text{neg}}$ incorrect responses have been collected (e.g., $K_{\text{pos}}=K_{\text{neg}}=8$).
\end{itemize}
Both strategies ensure that difficult prompts (low $p_i$) receive more samples, thereby realizing the goal of our theoretical framework. \textsc{Reinforce-Ada-Seq-balance} further emphasizes collecting failure cases, which intuitively corresponds to the behavior of GRPO-type weights that emphasizes prompts with pass rates near 0 or near 1. This design encourages gathering a more diverse set of training signals before a prompt is eventually deactivated.

\paragraph{Practical Implementation via Static Batches.}
A unique challenge of this sequential approach is that the total number of samples $N_i$ becomes a random variable, resulting in dynamic batch sizes that are incompatible with static GPU computational graphs. To make \textsc{Reinforce-Ada-Seq-pos/Balance} practical, we employ an Oversample-Downsample-Correct pipeline to convert these variable pools into a static training batch.


First, we leverage the variable pool of $N_i$ responses to compute a \textbf{high-fidelity global baseline}, $\hat{p}_i = N_{\text{pos}} / N_i$. Because we allocate larger budgets to difficult prompts, this estimate is now more likely to be non-zero and robust, far superior to statistics derived from small fixed groups.

Next, to create a static batch, we down-sample each prompt's pool to a fixed group of $n$ responses (e.g., $n=16$). We use a balanced sampling strategy (drawing $n/2$ positive and $n/2$ negative samples) to ensure signal diversity. While this down-sampling achieves engineering efficiency, it removes the implicit weight ($N_i$) generated by the sequential sampling process. Fortunately, the high-fidelity baseline $\hat{p}_i$ obtained in the first step allows us to recover the theoretical objective. Because $\hat{p}_i$ is stable and bounded away from zero (thanks to the sequential sampling process), we can safely return to the Explicit Weighting strategy. We simply re-introduce the weight by explicitly multiplying the gradient by $1/\hat{p}_i$.

The final gradient estimator for \textsc{Reinforce-Ada-Seq-positive/Balance} is thus:
$$\hat{g}(x_i) = \underbrace{\left( \frac{1}{\hat{p}_i} \right)}_{\text{Re-introduced Explicit Weight}} \cdot \underbrace{\left( \frac{1}{n} \sum_{j=1}^{n} \nabla \log \pi(a_j) \cdot (r_j - \hat{p}_i) \right)}_{\text{Static Group Gradient}}$$This pipeline enables the algorithm to operate with the throughput of static batches while mathematically optimizing the weighted non-linear objective.

\begin{table}[t]
\centering
\small
\renewcommand{\arraystretch}{1.3} 
\begin{tabular}{l p{9.5cm}} 
\toprule
\textbf{Method} & \textbf{Core Idea for Handling Zero-Variance} \\
\midrule

Uniform GRPO (small-$n$) &
No recovery mechanism. $\sigma(\{r\})=0$ forces advantage $A_i \to 0$, causing gradient vanish. \\

Passive Filtering \citep{yu2025dapo} &
Skips updates for groups with uniform rewards (data inefficient). \\

Large-$n$ GRPO &
Relies on large $n$ to render $\Pr[\sigma(\{r\})=0]$ negligible, approximating the true distribution. \\

\midrule

N-GRPO \citep{nan2025ngrpo} &
Augment the reward group with a constant (the maximum possible reward). \\

RL-ZVP \citep{le2025no} &
Assign advantages based on entropy information for prompts with uniform rewards. \\

\midrule 

\textbf{Reinforce-Ada-Est (Ours)} &
Estimates pass rate $\hat{p}$ (via EMA/Value Net) to estimate difficulty-based inference budget allocation ($n_i \propto 1/\sqrt{\hat{p}_i}$).
\\

\textbf{Reinforce-Ada-Seq (Ours)} & 
By design: always collects mixed outcomes before exit. \\
\bottomrule
\end{tabular}
\caption{Summary of methods for recovering signal under zero-variance groups.}
\label{tab:zero_var_methods}
\end{table}

\subsection{Comparison with the Dynamic Sampling in DAPO}

We now compare our framework with the dynamic sampling strategy proposed in DAPO \citep{yu2025dapo}. DAPO samples batches of prompts with a fixed group size $n$ and discards those that lack learning signals (i.e., groups with uniform rewards). To maintain a full training batch size $B$, DAPO repeatedly samples new batches of prompts until enough valid groups are collected.

While this approach effectively avoids wasted gradient computations, it fundamentally differs from our objective. Because DAPO uses a uniform group size for all prompts, difficult prompts (where $p(x)$ is low) are statistically destined to yield no signal and be discarded, and very easy prompts with $p(x)$ close to 1 are similarly likely to produce all-success groups that are also discarded. Consequently, DAPO neither resolves the signal-loss problem for hard prompts nor preserves signal for already-solvable ones, biasing training toward a narrow band of mid-difficulty problems and potentially limiting the model’s ability to improve at the frontier of its capabilities.

In contrast, our method reallocates compute within each prompt by assigning larger inference budgets to harder prompts. This helps the model obtain meaningful learning signals even on difficult prompts with low pass rates—something DAPO is unable to do. Finally, we note that DAPO’s accumulative-prompt dynamic sampling is not conflict with our adaptive sampling scheme. The two strategies can be combined to yield more stable and effective training.

\section{Experiment}

\subsection{Experimental Setup}

\paragraph{Models and Benchmarks.} We evaluate the generality of our framework across varying scales and architectures, using four foundation models: \textbf{Qwen2.5-Math} (1.5B and 7B), \textbf{Qwen3-4B-Instruct}, and \textbf{Llama-3.2-3B-Instruct}. For evaluation, we employ a comprehensive suite of mathematical reasoning benchmarks: \textbf{MATH500} \citep{hendrycks2021measuring}, \textbf{Minerva Math} \citep{lewkowycz2022solving}, and \textbf{OlympiadBench} \citep{he2024olympiadbench}. Furthermore, to test robustness on competition-level problems, we compile an \textbf{AIME-like} test set consisting of 230 problems sourced from recent competition: AIME24, AIME25, HMMT24, HMMT25, BRUMO25, AMC23, and CMIMC25 \citep{balunovic_srimatharena_2025}. All evaluations report the \textbf{Pass@1} accuracy averaged over 32 samples (Ave@32), generated with a temperature of $1.0$ and a max token limit of 4096. We will also include the analysis of reward-entropy trade-off.

\paragraph{Data Curation and Verifier.}
Training utilizes the standard subset of \textbf{OpenR1-Math-220k}\footnote{\url{https://huggingface.co/datasets/open-r1/OpenR1-Math-220k}}. We employ the \textbf{Math-Verify} tool \citep{Kydlicek_Math-Verify_Math_Verification} for automatic solution correctness verification. 
To ensure a high-quality training signal, we implement a standard preprocessing pipeline:
(1) Deduplication of prompts;
(2) Exclusion of prompts where reference solutions fail verification;
(3) \textbf{Difficulty-based Filtering:} Following \citet{yang2024qwen2}, we filter prompts based on empirical difficulty estimated by sampling 16 responses from the base model. We discard trivial prompts (average reward $> 0.375$) and effectively impossible prompts (0 correct responses). This helps us to focus the training distribution on problems of moderate difficulty rather than trivially easy or impossibly hard ones. This choice is motivated by the observation of \citet{openr1} that incorporating overly simple problems into the training set can degrade model performance.

\paragraph{RL Training Details.}
All experiments are conducted based on the \texttt{verl} framework \citep{sheng2024hybridflow}, where a Tinker-based\footnote{\url{https://github.com/thinking-machines-lab/tinker-cookbook}} implementation is also provided. Training is configured with a prompt batch size of 512. We use the AdamW optimizer with a fixed learning rate of $1 \times 10^{-6}$. To encourage exploration (with $10$-step warm-up), an entropy regularization term with coefficient $1 \times 10^{-4}$ is applied, while no KL penalty is introduced. Following \citet{yu2025dapo}, we adopt the clip-higher trick by setting the clipping range to $[0.2, 0.28]$. For all variants, we keep these standard RL hyperparameters (learning rate, batch size, entropy coefficient, optimizer, etc.) fixed. No tuning was performed for \textsc{Reinforce-Ada}. Each RL run is carried out for 600 training steps to obtain the final model.

\begin{table*}[htp]
    \centering
    \begin{adjustbox}{max width=\textwidth}
    \begin{tabular}{clccccc} 
    \toprule
        Model & Algorithm & \textbf{Math500} & \textbf{Minerva Math} & \textbf{Olympiad Bench} & \textbf{AIME-like} & \textbf{Weighted Average} \\
        \midrule
        \multirow{4}{*}{\textit{Qwen2.5-Math-1.5B}} 
        & GRPO-n4 & 74.2 & 34.4 & 38.4 & 16.2 & 45.3\\
             &   GRPO-n8 & 75.3 & 34.3 & 39.2 & 17.3 & 46.1\\
        & GRPO-n16 & 76.9 & 35.1 & 40.4 & 17.9 & 47.3\\
       & DAPO & 75.2 & 34.7 & 38.5 & 17.5 & 45.9 \\
               \cmidrule(l){2-7}
        & \textsc{Reinforce-Ada-Seq-pos} & 75.8 & 35.7 & 38.6 & 16.5 & 46.1\\
        & \textsc{Reinforce-Ada-Seq-balance} & 77.4 & 36.5 & 40.5 & 17.5 & \textbf{47.6}\\
        & \textsc{Reinforce-Ada-Est}
        & 76.1 & 34.8 & 39.1 &18.5 &  46.5 
        \\
            \midrule
        \multirow{4}{*}{\textit{Qwen2.5-Math-7B}} 
        & GRPO & 82.2 & 44.7 & 45.6 & 23.2 & 53.3\\
        & \textsc{Reinforce-Ada-Seq-pos} & 82.7 & 45.1 & 46.7 & 23.7 & 54.2\\
        & \textsc{Reinforce-Ada-Seq-balance} & 84.0 & 45.2 & 47.1 & 23.7  & \textbf{54.6}\\
        & \textsc{Reinforce-Ada-Est}
        & 83.6 & 42.1& 46.6 & 24.1 &53.7 
        \\
           \midrule
        \multirow{4}{*}{\textit{Llama-3.2-3B-instruct}} 
        & GRPO & 51.7  & 20.5 & 20.4 & 7.2 & 27.9\\
        & \textsc{Reinforce-Ada-Seq-pos} & 52.6 & 22.2 & 21.0 & 7.5 & 28.8 \\
        & \textsc{Reinforce-Ada-Seq-balance} & 53.2 & 22.4 & 21.2 & 8.0 & \textbf{29.1} \\
        \midrule
        \multirow{4}{*}{\textit{Qwen3-4B-instruct}} 
        & GRPO & 90.4 & 51.2 & 64.9 & 38.5 & 66.5 \\
        & \textsc{Reinforce-Ada-Seq-pos} & 91.6 & 50.4 & 66.3 & 38.8 & 67.4 \\
        & \textsc{Reinforce-Ada-Seq-balance} & 91.7 & 53.0 & 65.7 & 38.8 & \textbf{67.6}  \\
    \bottomrule
    \end{tabular}
    \end{adjustbox}
        \caption{Performance comparison of GRPO and \textsc{Reinforce-Ada}. 
        We report average@32 accuracy with a sampling temperature of 1.0 and a maximum generation length of 4096 tokens. The weighted average score is computed according to the number of prompts in each benchmark. }    \label{tab:main_res2}
\end{table*}

\begin{figure}[htp]
    \centering
    \includegraphics[width=0.75\textwidth]{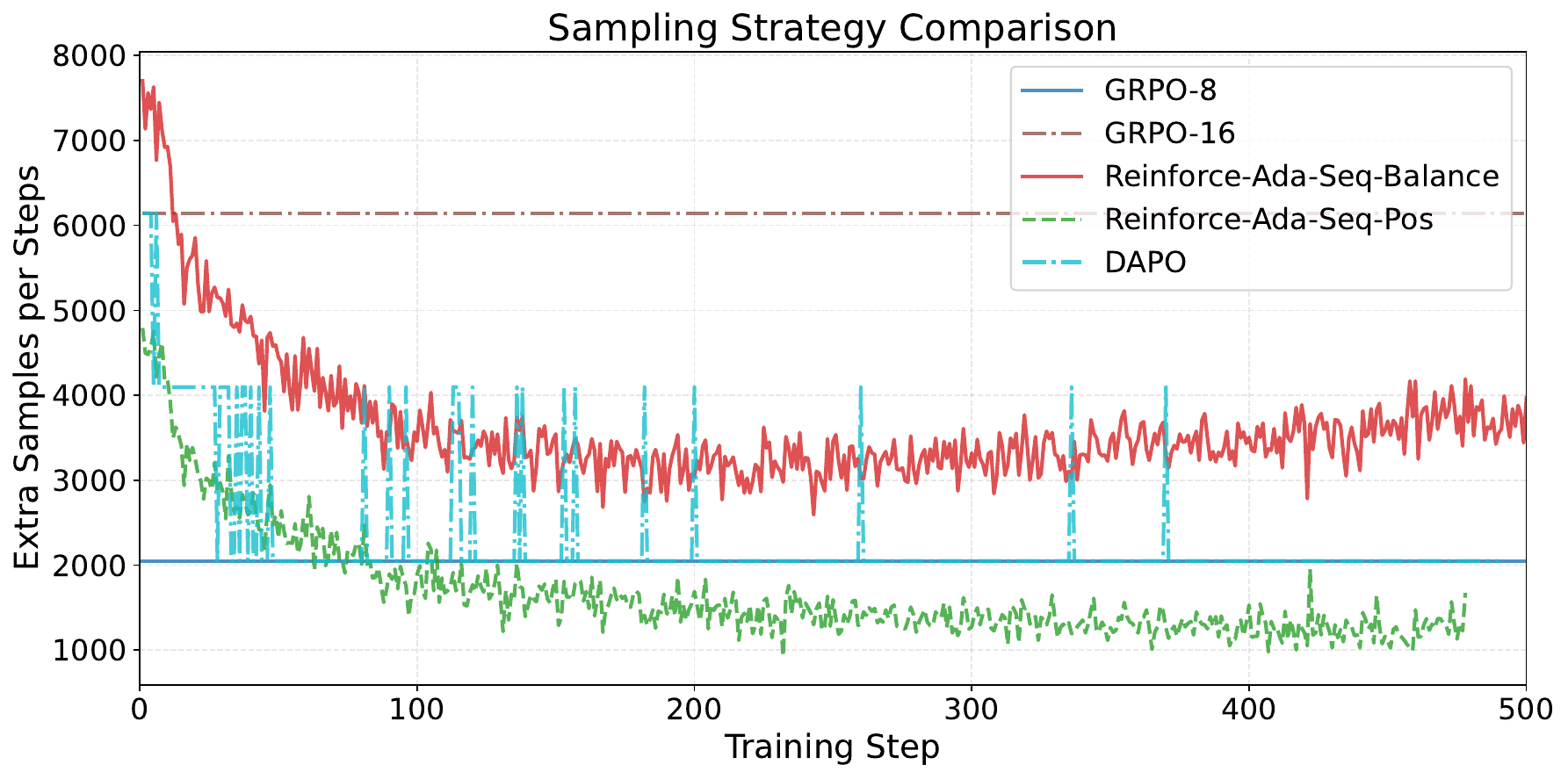}      \includegraphics[width=0.32\textwidth]{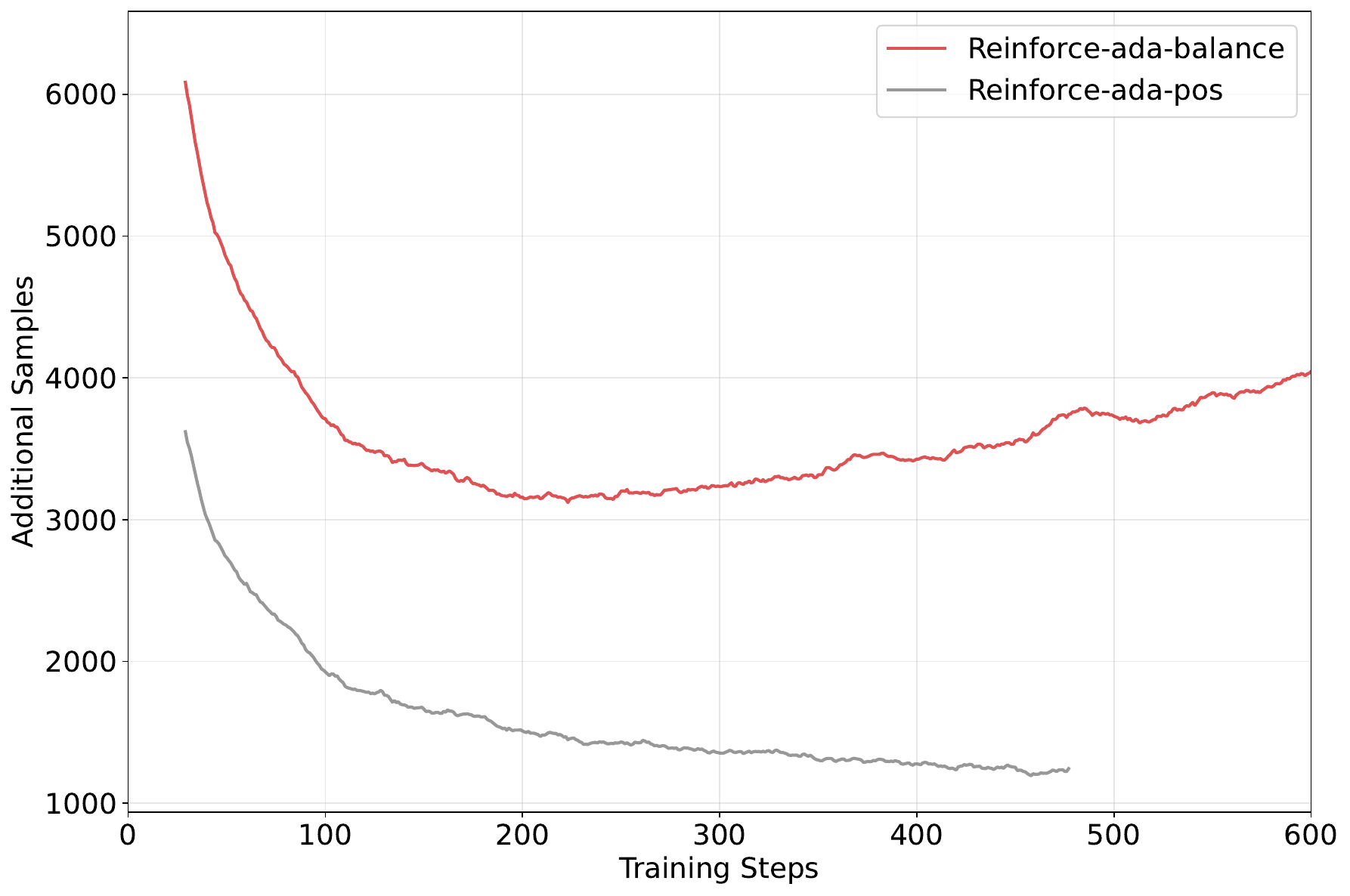}  
    \includegraphics[width=0.32\textwidth]{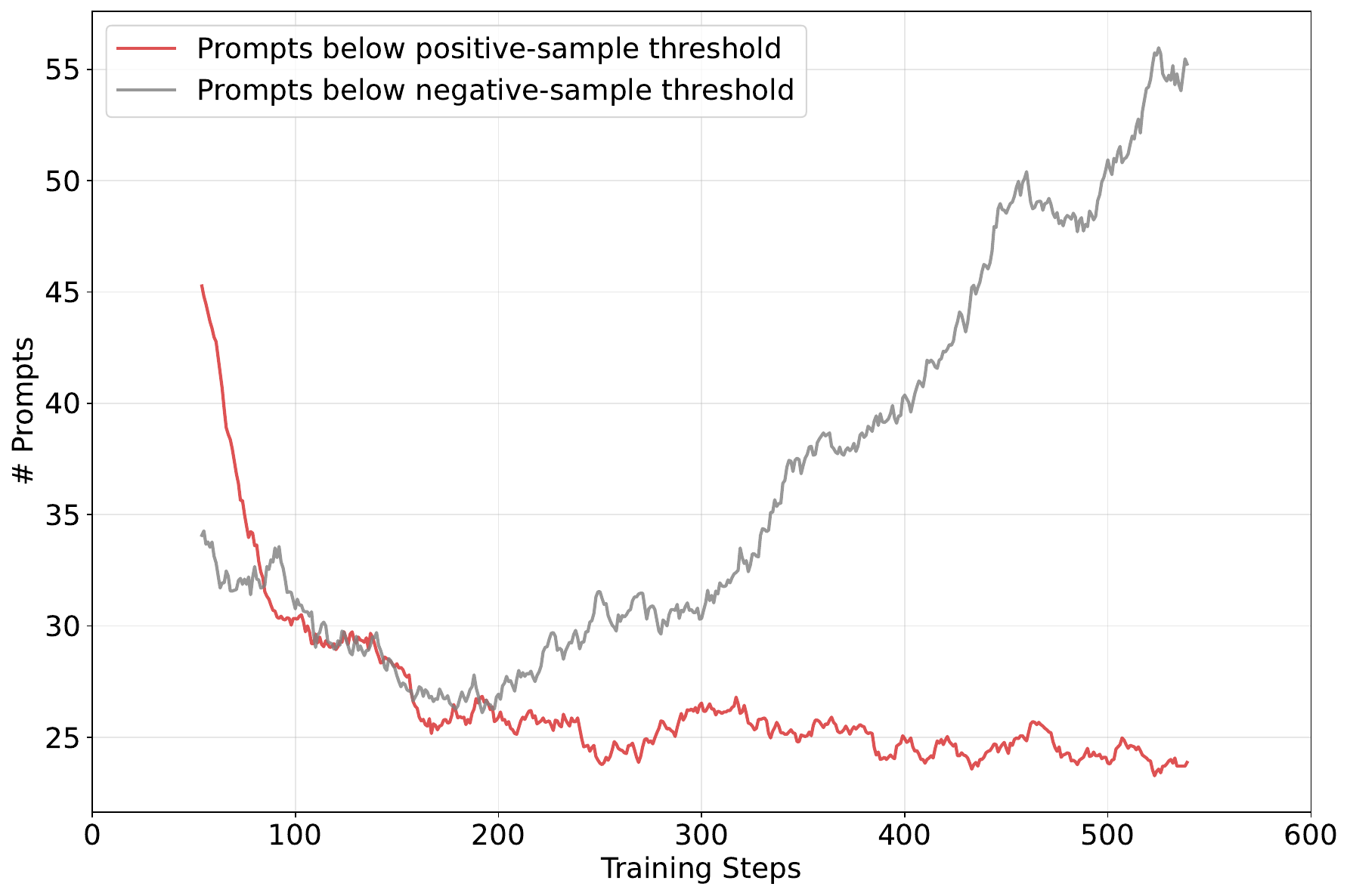}
    \includegraphics[width=0.32\textwidth]{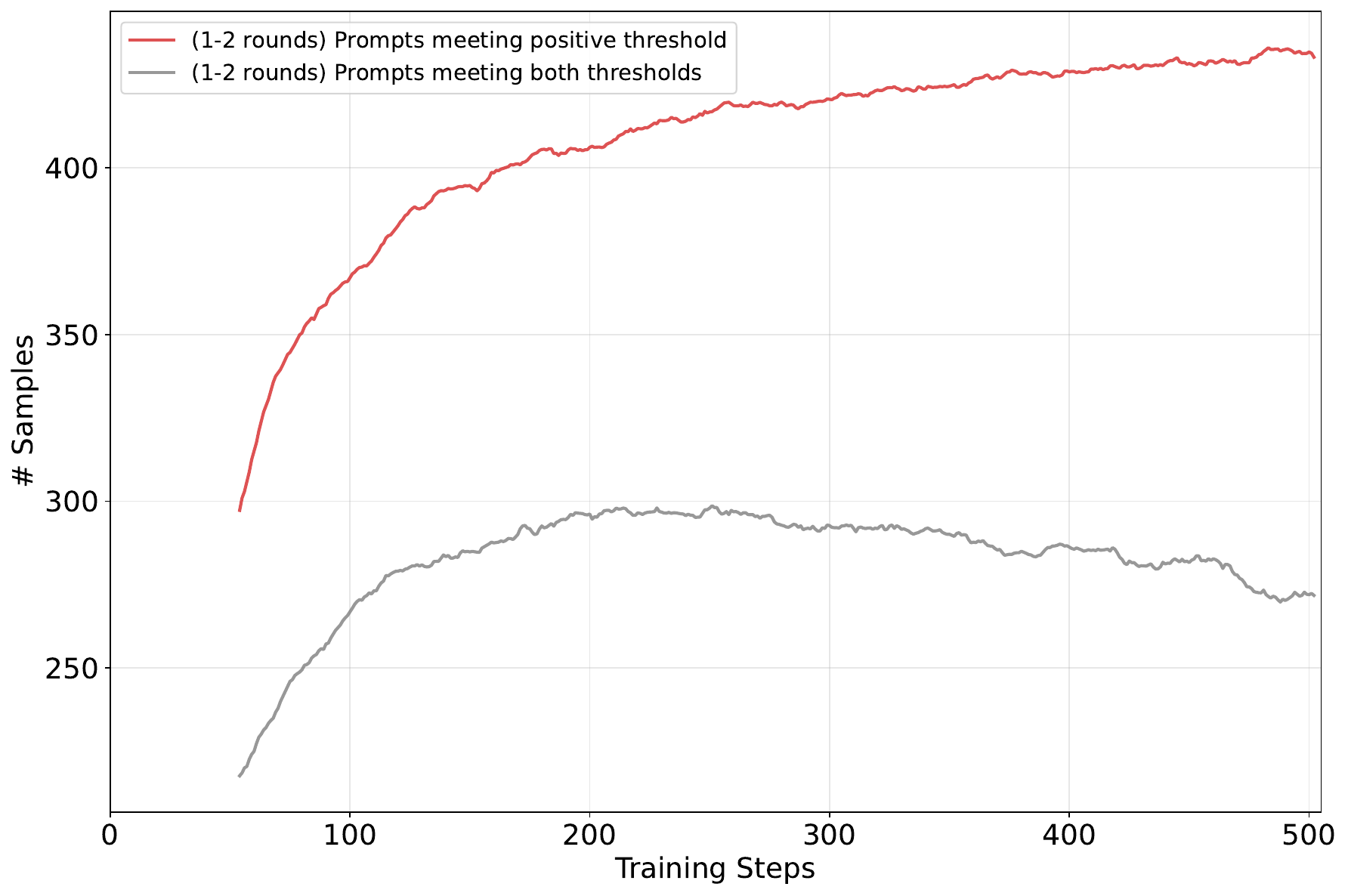}
\caption{First row: Sampling dynamics of different training strategies using the Qwen2.5-Math-1.5B model. We omit \textsc{Reinforce-Ada-Est} since its sampling cost matches that of GRPO-8. Second row: Sampling dynamics with the Qwen2.5-Math-1.5B model. Left: additional samples generated in later rounds compared to standard GRPO. Middle: number of prompts that remain active after multi-round adaptive sampling with the \textsc{Reinforce-Ada-Seq-balance} variant. Right: number of prompts that satisfy the stopping criteria within the first two rounds with the \textsc{Reinforce-Ada-Seq-balance} variant. All curves are smoothed using a moving average with a window size of $20$.
}
        \label{fig:sampling_dynamic} 
\end{figure}

\paragraph{Sampling Configuration.} The downsampling mechanism and random samples of \textsc{Reinforce-Ada-Seq} make the methods not directly comparable. We standardize the \textbf{update group size} (the number of samples used for the backward pass) to $n=4$ for \textsc{Reinforce-Ada-Seq} and baseline GRPO, but we will also include the GRPO baselines with larger group size $8$ and $16$. We refer readers to Figure~\ref{fig:sampling_dynamic} to get a sense of the sampling dynamic of different methods. Specifically, we will consider the following competitors:
\begin{itemize} \item \textbf{GRPO Baselines:} We compare against GRPO with fixed group sizes $n \in \{4, 8, 16\}$.
\item \textbf{Reinforce-Ada-Seq:} The algorithm adaptively samples up to $N_{\max}=32$ responses per prompt but downsamples the collected pool to a fixed group size of $n=4$ for the gradient update. This ensures that the training cost (backward pass) remains identical to GRPO-n4, with overhead limited only to inference. 
\item \textbf{Reinforce-Ada-Est:} The total sampling budget is allocated proportionally to $1/\sqrt{\hat{p}_i}$. The choice of hyper-parameter $8$ is to roughly match the sampling cost of the \textsc{Reinforce-Ada-Seq}
\end{itemize}
In general, \textsc{Reinforce-Ada-Est} is most directly comparable to GRPO-$n8$, as they consume approximately the same number of forward and backward samples. In contrast, \textsc{Reinforce-Ada-Seq} involves inherent stochasticity in its adaptive sampling process, so only coarse-grained comparisons are possible; nevertheless, our experiments consistently show clear and robust trends across different models.

\subsection{Main Result}

\begin{figure}[htp]
    \centering
    \includegraphics[width=0.49\textwidth]{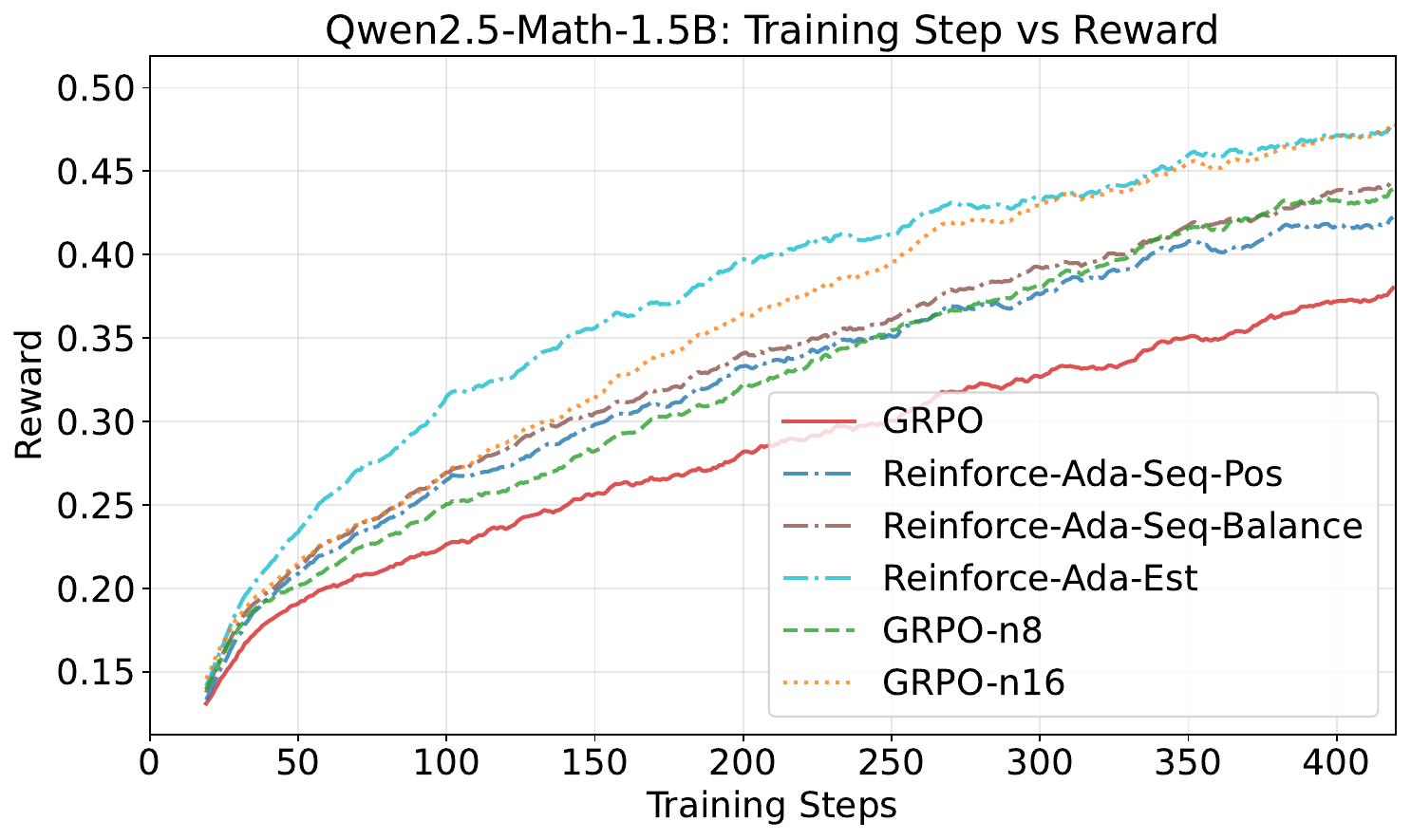}  
    \includegraphics[width=0.49\textwidth]{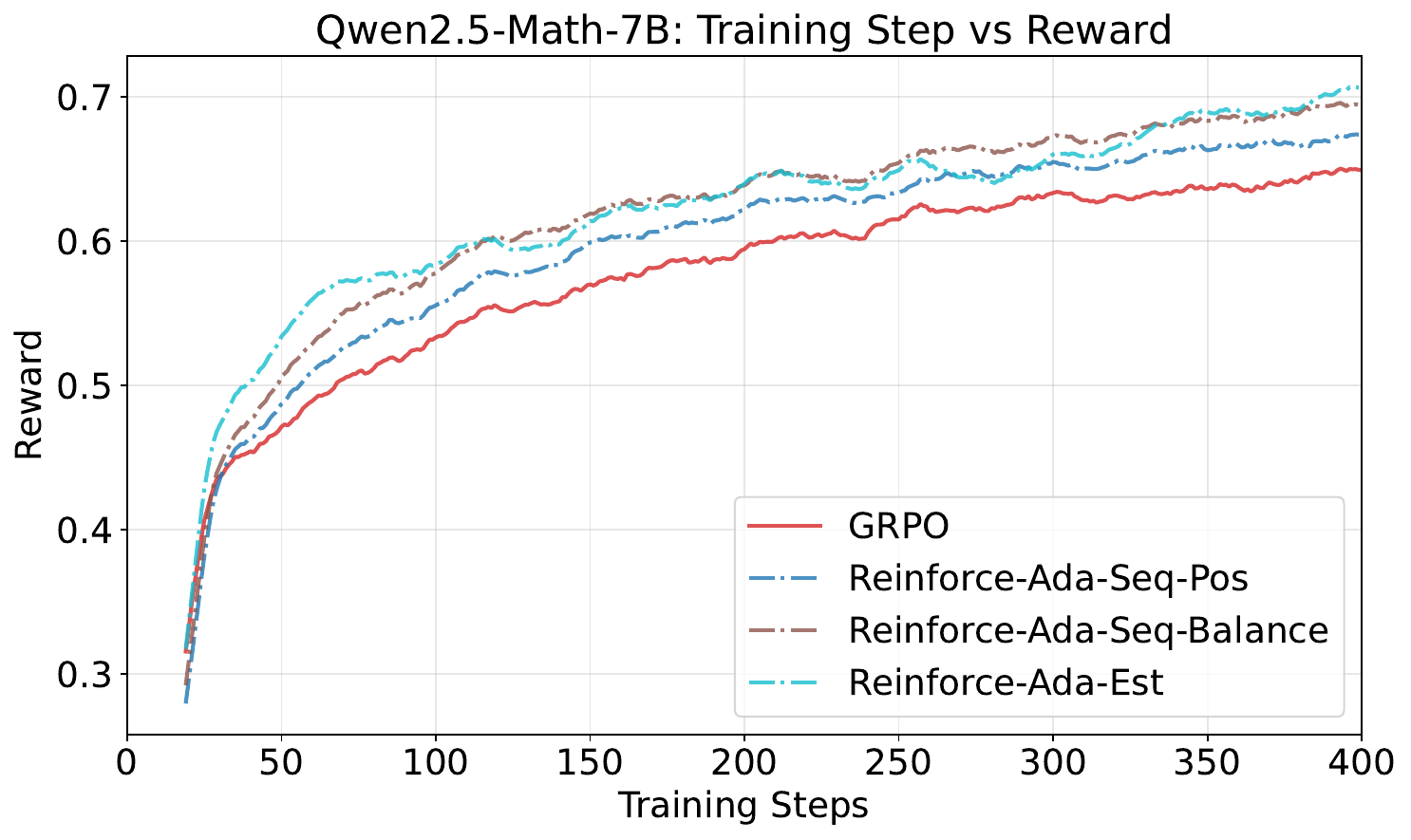}
    \includegraphics[width=0.49\textwidth]{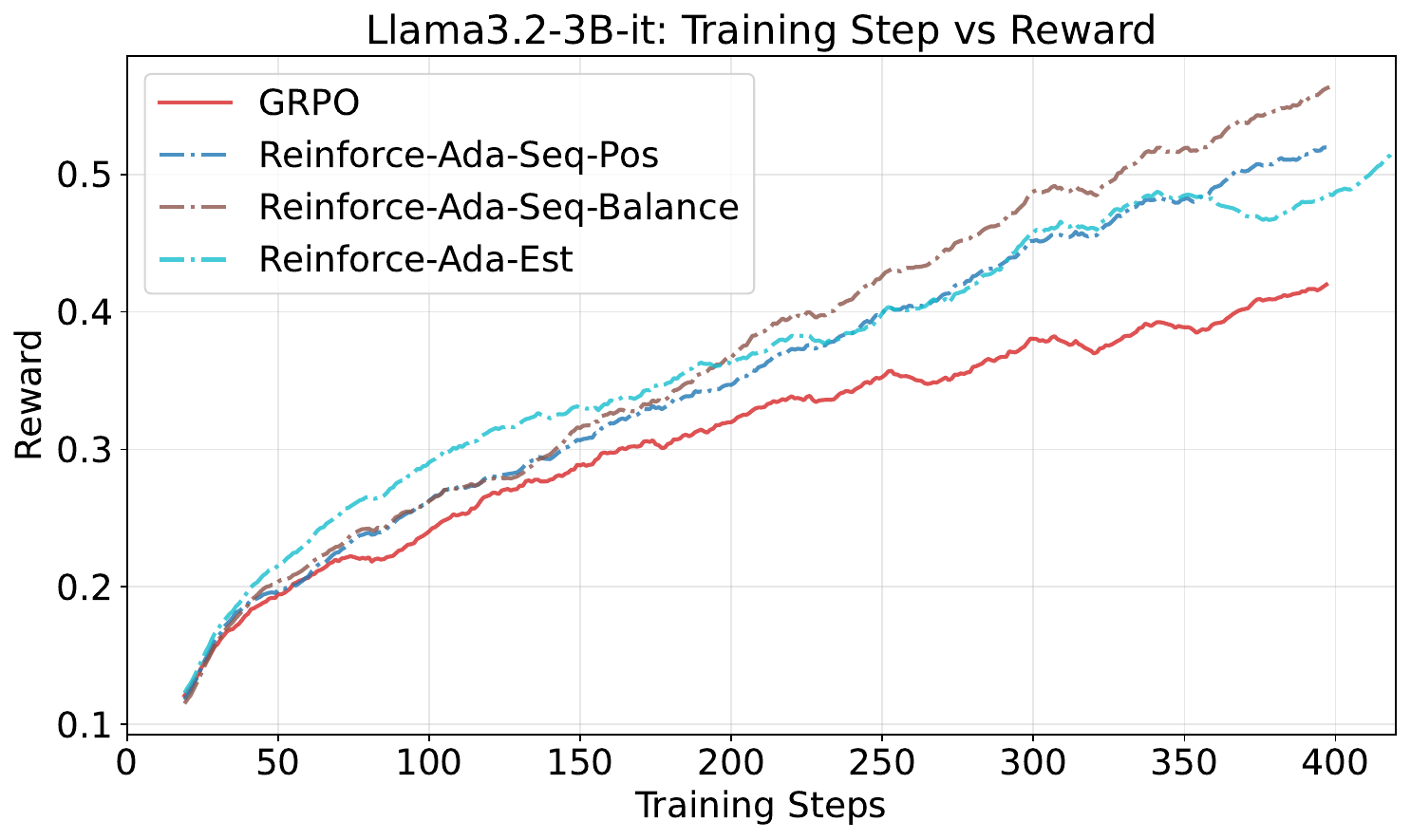}
    \includegraphics[width=0.49\textwidth]{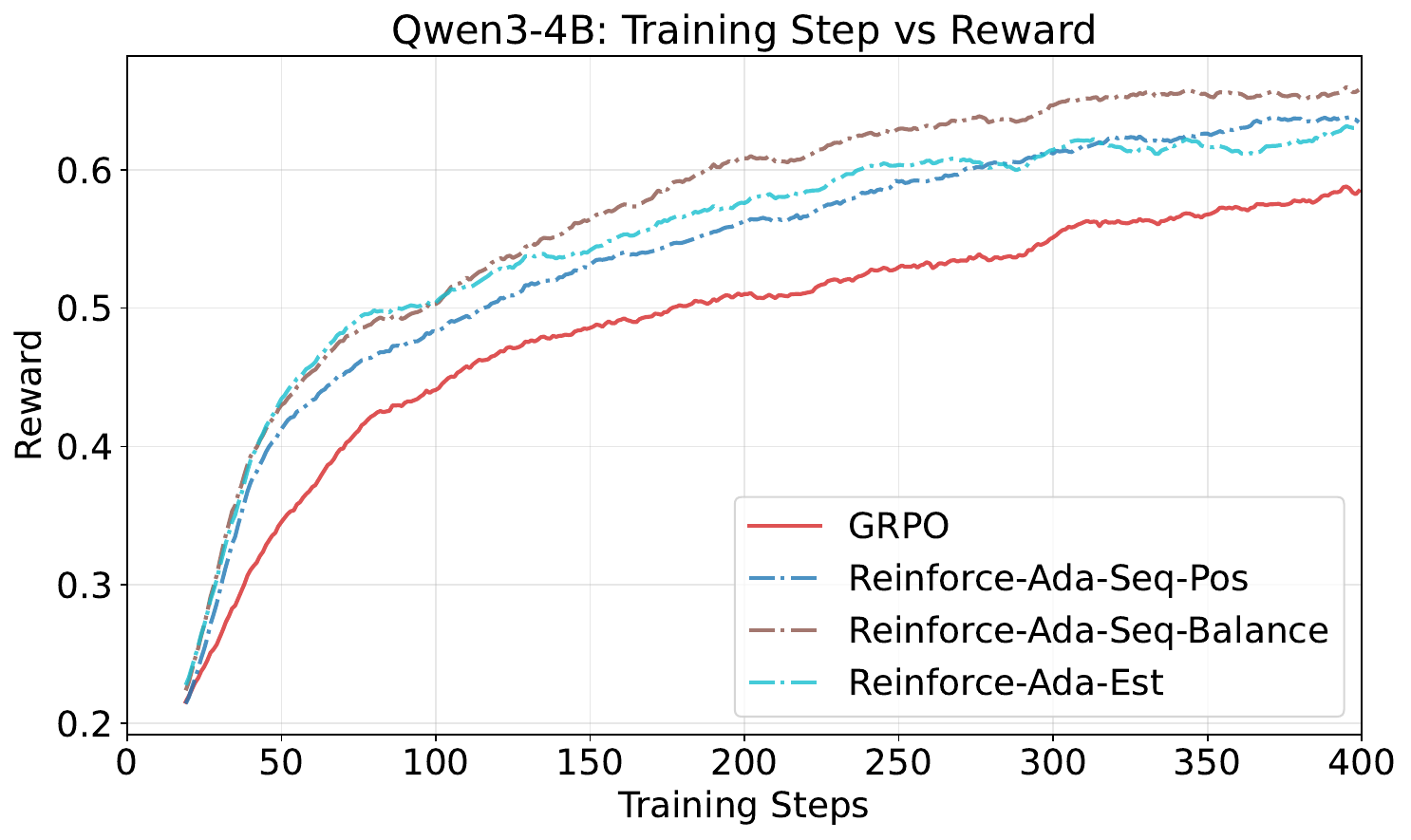}
\caption{Training reward vs.\ steps for GRPO and \textsc{Reinforce-Ada} across backbones:
Qwen2.5-Math-1.5B, Qwen2.5-Math-7B, and Llama-3.2-3B-it, Qwen3-4B. Curves are smoothed with a 20-step moving average. In all cases, \textsc{Reinforce-Ada} learns faster and reaches a higher reward than GRPO, with the \textsc{Balance} variant typically achieving the highest asymptote. 
}
        \label{fig:training_reward} 
\end{figure}

\paragraph{Adaptive Allocation vs. Uniform Scaling.}
Table~\ref{tab:main_res2} summarizes the performance across all benchmarks and Figure~\ref{fig:training_reward} presents the training reward curves. We first notice that standard GRPO shows monotonic improvement as group size increases from $n=4$ to $n=16$ (e.g., $45.3\% \to 47.3\%$ on Qwen2.5-Math-1.5B). This confirms that signal loss is an artifact of undersampling. However, this comes at a linear cost. \textsc{Reinforce-Ada} demonstrates a more efficient pathway to scale up RL training. Instead of uniformly increasing the budget for all prompts, our method dynamically directs compute to the learning frontier.
Crucially, \textsc{Seq-Balance} successfully matches or even exceeds the performance of the expensive GRPO-n16 (e.g., $47.6\%$ vs. $47.3\%$ on Qwen2.5-Math-1.5B) while operating with a significantly lower inference budget (comparable to $n \approx 8-10$) and lower training cost ($n=4$). This indicates that \textsc{Reinforce-Ada} can capture the signal-quality benefits of large-scale sampling without incurring the full computational penalty.

\paragraph{Comparison with Passive Filtering (DAPO).}
While DAPO improves over the naive GRPO-n4 ($45.9\%$ vs $45.3\%$) by filtering out zero-signal groups, it generally lags behind our adaptive methods. This supports our hypothesis that \textit{passive} filtering is insufficient for difficult reasoning tasks; instead, \textit{active} investment of compute is required to recover the latent learning signals in hard prompts. In Appendix~\ref{sec:diff_level}, we will provide a more detailed analysis of the accuracies on prompts with different difficulty levels, which further validates our intuition.

\paragraph{Comparison among different adaptive sampling variants.} Comparing the adaptive variants, we observe that \textsc{Seq-Balance} consistently outperforms \textsc{Seq-Pos}, with the gap widening in later training stages.
As the policy improves, positive responses become abundant, causing \textsc{Seq-Pos} to exit early and revert to a near-uniform sampling regime. However, negative responses (errors) become rare for these improved models. \textsc{Seq-Balance} forces the collection of these "hard negatives," ensuring that the variance of the gradient estimator remains non-zero. This sustained signal quality explains its higher asymptotic performance and robustness across different model families. \textsc{Ada-Seq} and \textsc{Ada-Est} are not directly comparable as their inference and training costs are not identical. We also observe a mixed result across different mdoels. For instance, with Qwen2.5-Math-1.5B, \textsc{Ada-Est} is the best one. But for Llama-3.2-3B-it, we notice that the \textsc{Ada-Est} roughly matches that of \textsc{Ada-Seq-Pos} but is behind the \textsc{Ada-Seq-Balance}. 

Finally, we emphasize that the additional computational overhead of \textsc{Reinforce-Ada-Seq} stems \emph{only} from the inference (generation) stage. Because we uniformly down-sample the collected trajectories to a fixed group size for the policy update, the backward pass (training cost) remains identical to standard GRPO-n4. In contrast, GRPO-n8, GRPO-n16, and \textsc{Ada-Est} will incur a much larger training cost than other methods since the train on more samples per iteration.

\subsection{Analysis of Efficiency and Learning Dynamics}
\label{sec:dynamic_and_ablation}

\paragraph{Computation vs. Performance Sweet Spot.}
We would like to quantify the efficiency gains and analyze the trade-off between inference cost (average samples generated per prompt) and final accuracy. Figure~\ref{fig:sampling_dynamic} summarizes the \emph{inference cost}\footnote{We remark that GRPO-n16 uses more samples per iteration, resulting in a higher backward-pass cost.} measured as the average number of generated samples per iteration. The ranking is 
\begin{itemize}
    \item \textbf{Inference Cost Ranking:} GRPO-n4 $<$ \textsc{Seq-Pos} $<$ \textsc{Seq-Balance} $<$ GRPO-n16.
    \item \textbf{Performance Ranking:} GRPO-n4 $<$ \textsc{Seq-Pos} $<$ GRPO-n16 $\approx$ \textsc{Seq-Balance} $<$ \textsc{Ada-Est}.
\end{itemize}
This comparison highlights the efficiency of our approach. \textsc{Ada-Est} achieves the high performance of the expensive \textsc{GRPO-n16} baseline but requires significantly less compute (lower than GRPO-n16). By allocating the budget adaptively, we can capture the benefits of scaling up group size (recovering lost signals) without the waste of generating many samples for easy prompts.

\begin{table}[htp]
    \centering
    \begin{tabular}{lccc}
        \toprule
        \textbf{Model} & \textbf{Algorithm} & \textbf{Avg. Step Time (s)} & \textbf{Relative Cost} \\
        \midrule
        \multirow{4}{*}{Qwen2.5-Math-1.5B} 
        & GRPO & 102 & $1.0\times$ \\
        & \textsc{Reinforce-Ada-Seq-pos} & 228 & $2.2\times$ \\
        & \textsc{Reinforce-Ada-Seq-balance} & 290 & $2.8\times$ \\ 
        & \textsc{Reinforce-Ada-Est} & $128$ & $1.3\times$
        \\
        \midrule
        \multirow{4}{*}{Qwen2.5-Math-7B} 
        & GRPO & 236 & $1.0\times$ \\
        & \textsc{Reinforce-Ada-Seq-pos} & 333 & $1.41\times$ \\
        & \textsc{Reinforce-Ada-Seq-balance} & 375 & $1.59\times$ \\ 
        & \textsc{Reinforce-Ada-Est} &382&$1.62\times$\\
        \bottomrule
    \end{tabular}
    \caption{Average step time (wall-clock seconds per update) of GRPO vs. \textsc{Reinforce-Ada} on 8$\times$H100 (1.5B) and 8$\times$A100 (7B). 
    Relative cost is normalized against GRPO for the same model.} 
    \label{tab:step_time}
\end{table}

\paragraph{Wall-Clock Efficiency.} We quantify the wall-clock cost of our adaptive sampling. We report the average per-step time with \texttt{verl} implementation in Table~\ref{tab:step_time}. Here 1.5B model is with 8$\times$NVIDIA H100, and 7B model is with 8$\times$NVIDIA A100. Naturally, \textsc{Ada-Seq} incurs a higher inference cost compared to the baseline GRPO ($n=4$) because it generates additional responses to hunt for sparse signals\footnote{Currently, the multi-sample generation in \textsc{Ada-Seq} is implemented synchronously, meaning that each optimization step is bottlenecked by the slowest completion among the sampled responses. We expect that an asynchronous implementation would be much faster.}. For the 1.5B model, \textsc{Seq-pos} and \textsc{Seq-balance} increase the step time by approximately $2.2\times$ and $2.8\times$, respectively. In comparison, the relative overhead of \textsc{Reinforce-Ada} is smaller on the 7B model than on the 1.5B model. From Figure~\ref{fig:training_reward}, the 7B policy exhibits a sharp early jump in training reward, which means that most prompts quickly satisfy the positive-sample stopping criterion. Consequently, adaptive sampling requires far fewer additional rollouts on 7B than on 1.5B. Second, the actor update dominates the per-step time on the 7B model (85s out of 236s), whereas it is only a small portion on the 1.5B model (15s out of 102s). As a result, the extra inference introduced by adaptive sampling has a much smaller impact on the overall step time for the 7B model.

\paragraph{The Sampling Dynamic of \textsc{Ada-Seq}.}
Figure~\ref{fig:sampling_dynamic} provides a deep dive into how \textsc{Ada-Seq} allocates its budget over time, where the experiments are with Qwen2.5-Math-1.5B:
\begin{itemize}
    \item \textbf{Early Training (Exploration Phase):} The cost is high as the model struggles with difficult prompts. The algorithm automatically allocates more samples to find the sparse positive solutions needed to initiate learning.
    \item \textbf{Mid Training (Consolidation Phase):} As the model improves, positive samples become easier to find. The sampling cost for \textsc{Seq-Pos} drops monotonically.
    \item \textbf{Late Training (Refinement Phase):} Crucially, \textsc{Seq-Balance} exhibits a distinct ``U-shaped'' cost curve. After $\sim$300 steps, the cost rises again. This phenomenon occurs because the model becomes highly proficient, making \textit{negative} samples (failures) rare. To maintain a valid gradient signal (variance $>0$), the algorithm actively hunts for these hard negatives. This adaptivity prevents the ``all-correct'' signal collapse that plagues standard GRPO in later stages.
\end{itemize}

\subsection{Entropy Dynamic and Pass@k}
Beyond simple accuracy, we evaluate the quality of the learned policy through the lens of the Reward-Entropy trade-off. A known challenge in evaluating the reasoning ability of LLMs is on the trade-off between reward (accuracy) and entropy (generation diversity). Previous work has argued that post-training can trade this uncertainty for higher rewards in a predictable manner \citep{cui2025entropy}. Moreover, prior work shows that standard GRPO often collapses to a low-entropy policy early in training, reducing reasoning diversity thus a worse pass@k for large k \citep{shao2024deepseekmath, yue2025does}.

\begin{figure}[htp]
    \centering
    \includegraphics[width=0.47\textwidth]{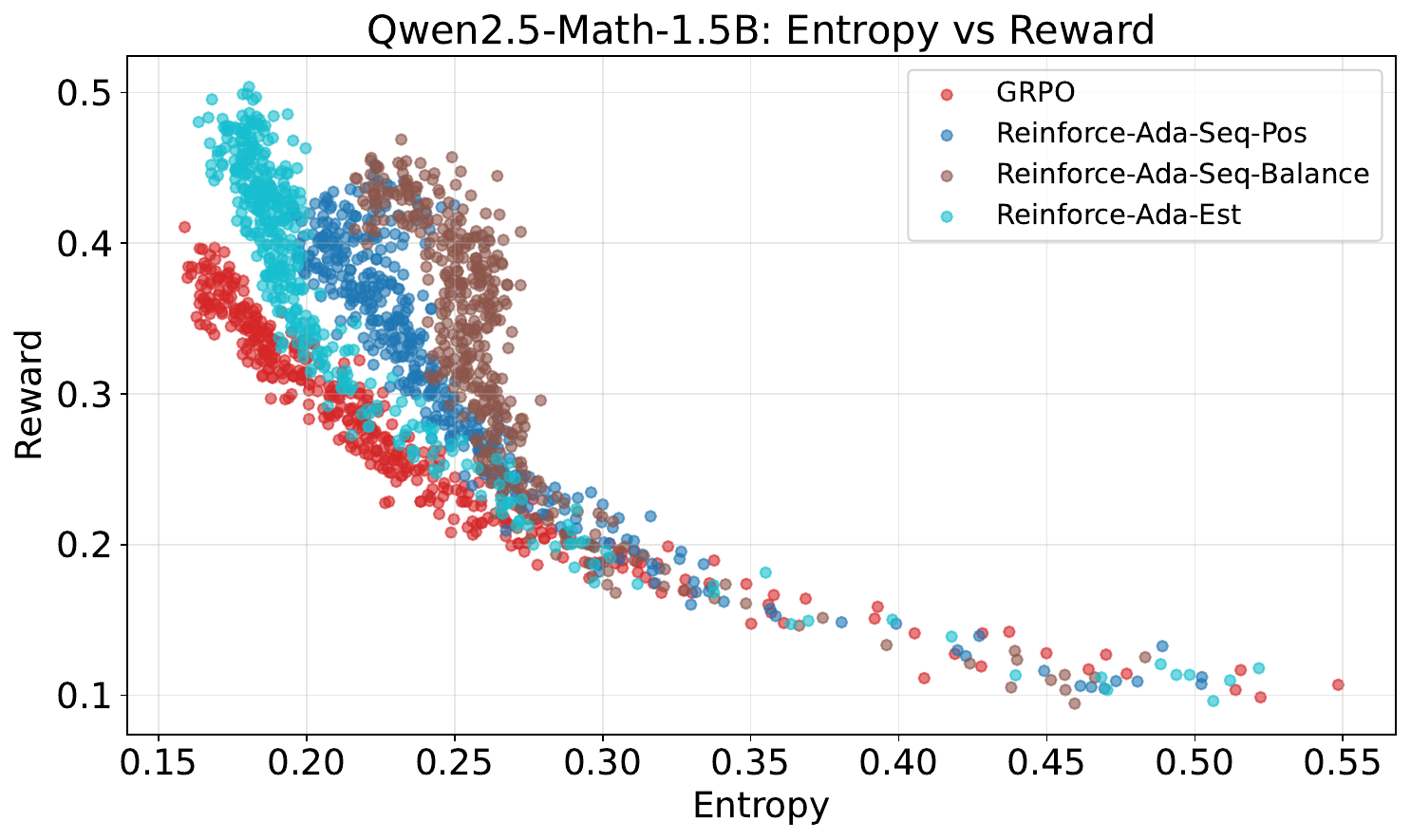}  
    \includegraphics[width=0.47\textwidth]{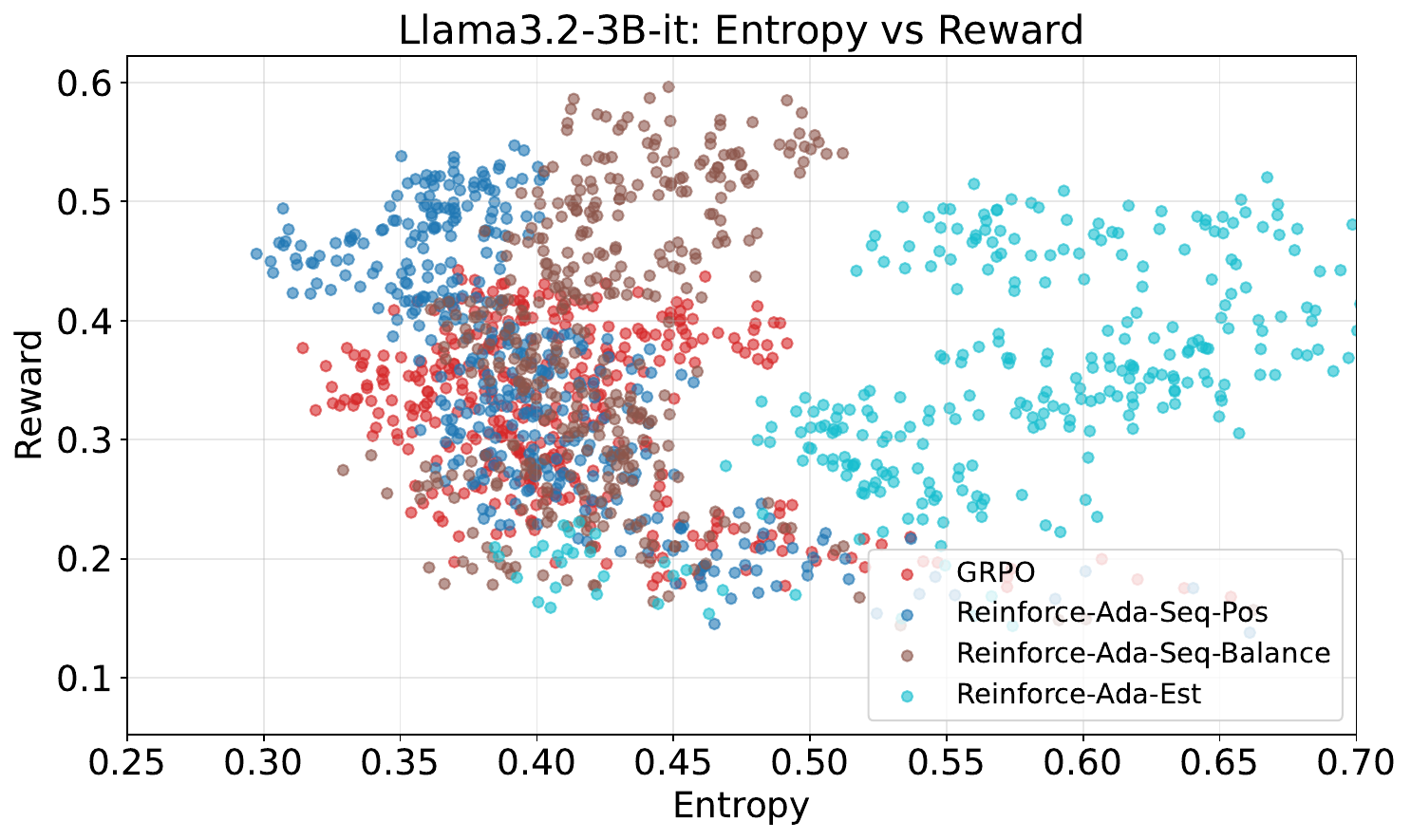}  
    \includegraphics[width=0.47\textwidth]{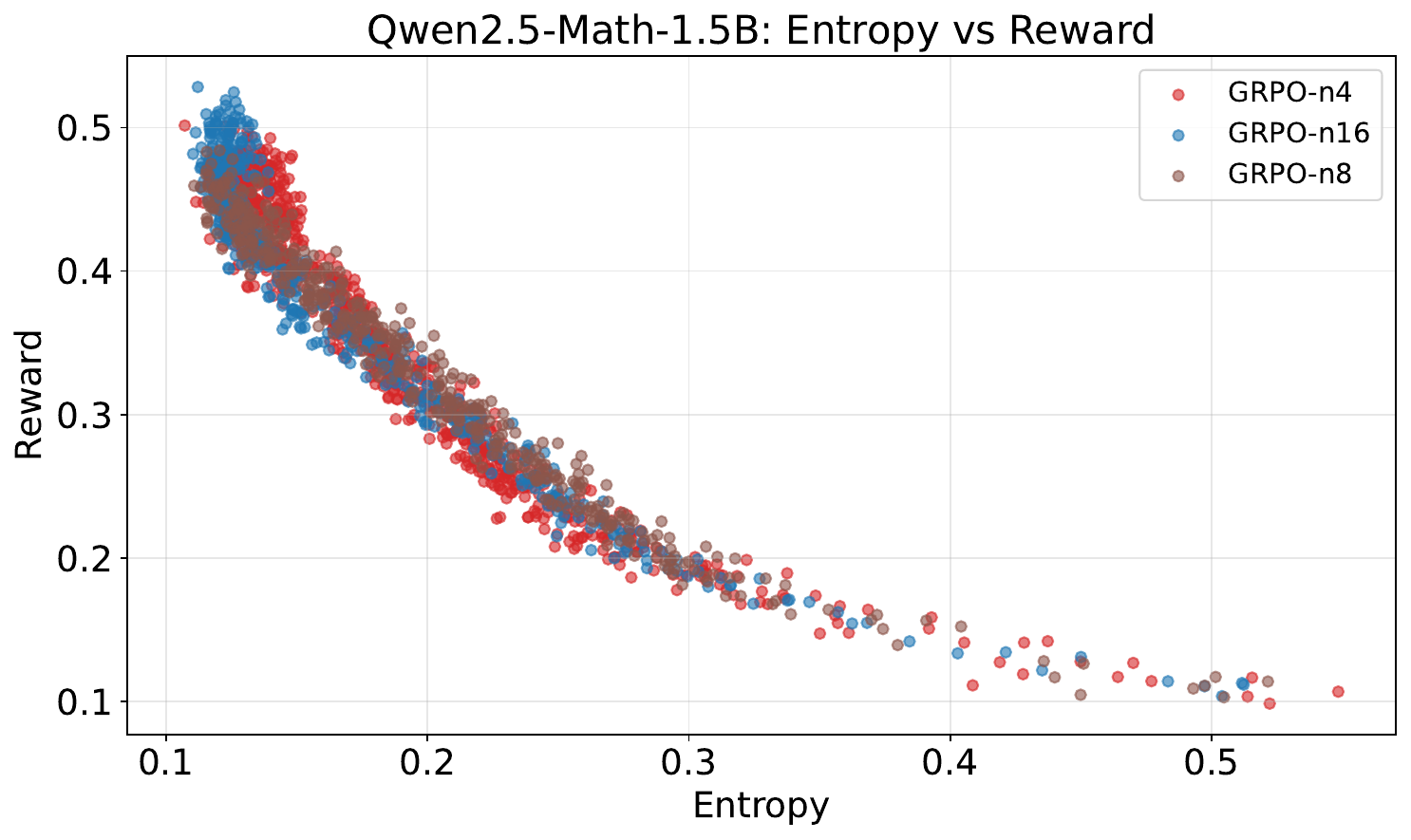} 
\includegraphics[width=0.47\textwidth]{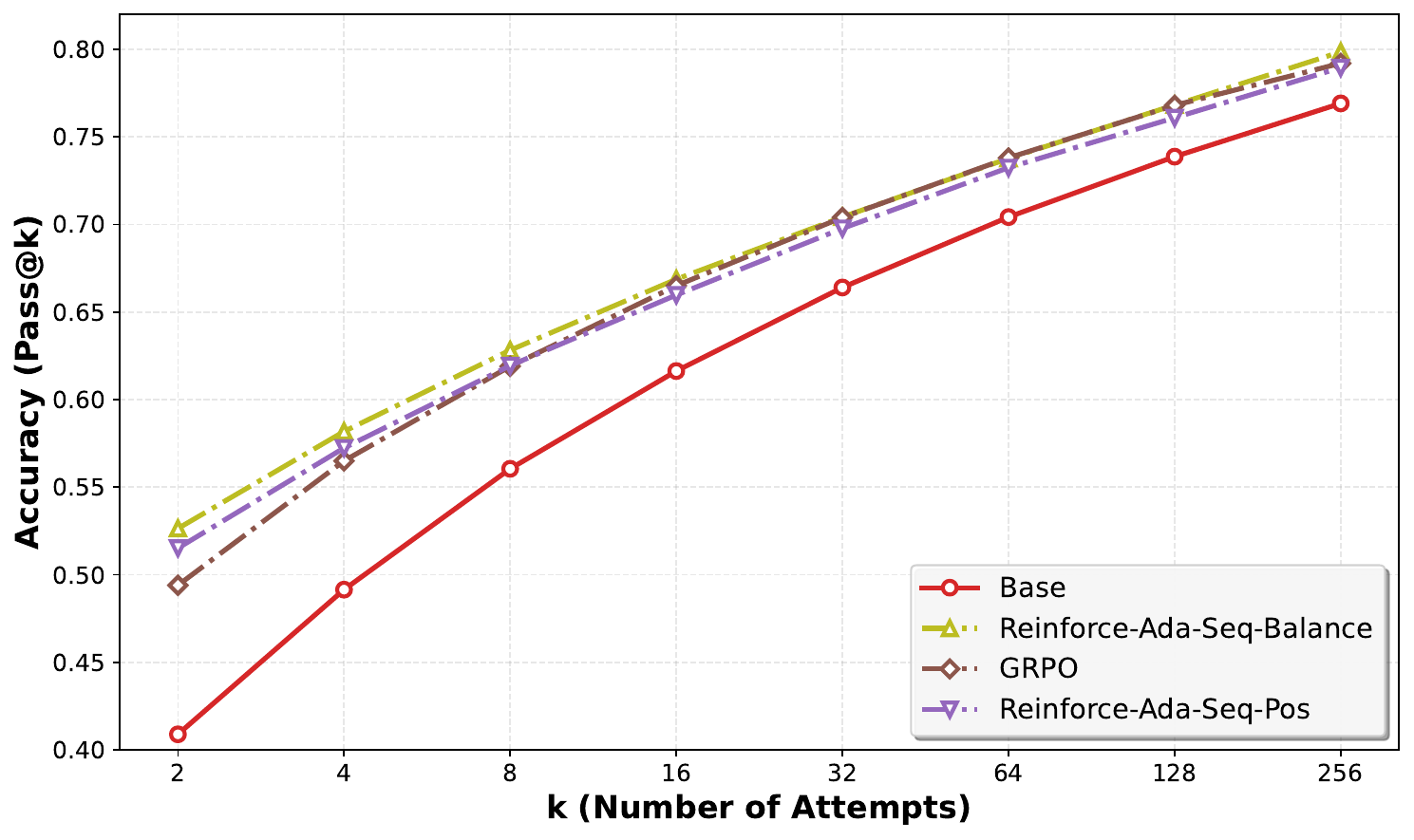}  
        \caption{{Reward–entropy trade-off (left and mid) and Pass@k (right) on the test benchmarks. \textsc{Reinforce-Ada} shifts the frontier outward: higher reward at fixed entropy and higher entropy at fixed reward and converts this into stronger Pass@k at small, practical budgets ($k\le8$), with \textsc{Reinforce-Ada-Seq-balance} typically best.}}\label{fig:entropy_passk} 
\end{figure}

An ideal algorithm should improve rewards without causing the policy to become overly deterministic (entropy collapse).  To provide a more comprehensive evaluation, we analyze this trade-off in this part and present the results in Figure~\ref{fig:entropy_passk}.
\begin{enumerate}[(i)]
    \item \textbf{Reward-entropy frontier.} While the precise entropy dynamics can vary depending on the foundation model, \textsc{Reinforce-Ada} consistently achieves a comparable or superior reward-entropy curve. Specifically, from (2-1) of Figure~\ref{fig:entropy_passk}, we notice that increasing the group size of GRPO cannot improve the reward-entropy trade-off. In comparison, on Qwen2.5-Math-1.5B (1-1), GRPO concentrates mass early (low entropy, narrow cloud) and achieves lower reward for a given entropy. All \textsc{Reinforce-Ada} variants shift the frontier outward and \textsc{Reinforce-Ada-Seq-balance} lies furthest, at equal reward it sustains higher entropy, and at equal entropy it achieves higher reward. On Llama-3.2-3B-it (1-2), the base policy starts with a higher entropy floor, so the separation among different methods is smaller. But the \textsc{Ada-Seq} variants still achieve a competitive frontier than GRPO and \textsc{Ada-Est} performs even better. We hypothesize that this is because the adaptive sampling pushes the model to explore towards the prompts they are uncertain, thus preserving the entropy. We also observe that \textsc{Reinforce-Ada-Seq-balance} $>$ \textsc{Reinforce-Ada-Seq-pos}.
    We attribute this to the fact that exposure to negative signals discourages the model from becoming overconfident in a single solution path, thus preserving valuable policy diversity. 
    \item \textbf{Pass@k behavior.} An related observation is that the moderate policy entropy typically converts to an improved pass@k compared to the base model for a wide range of k. Specifically, panel (2-2) shows that all RL methods dominate the base across $k$, but the largest, most practical gains appear at small budgets ($k\le8$). In this regime, \textsc{Reinforce-Ada} yields the highest Pass@k; the advantage narrows as $k$ grows (diminishing-returns regime), where \textsc{Reinforce-Ada-Seq-balance} typically remains marginally best. The pattern implies two complementary effects: (a) improved top-1 quality (higher reward at given entropy), and (b) retained diversity among high-scoring modes (shallower saturation with $k$). Together, adaptive sampling moves the reward–entropy curve outward and converts that shift into higher Pass@k at realistic attempt budgets, with \textsc{Reinforce-Ada-Seq-balance} offering the most stable trade-off.
\end{enumerate}

\section{Related Work}

\paragraph{Data filtering and selection in online RL training for LLMs.}
Our work is related to the growing body of literature on data selection and filtering for online reinforcement learning with LLMs, which further dates back to the RLHF studies \citep{zhang2024policy, xiong2023iterative, dong2024rlhf, shi2024crucial, feng2025pilaf}. In the context of RLVR, some methods employ an oversample-then-downsample strategy: they first generate a large, uniform set of responses for each prompt and then select a subset based on specific criteria. \citet{xu2025not} propose downsampling to maximize reward variance within the group, while \citet{ye2025beyond} use process rewards to select positive samples, mitigating issues with falsely correct responses. \citet{xue2025simpletir} study the data filtering in the context of tool-integrated reasoning, where they find that the trajectories with invalid tool calling will significantly hurt the training stability. \citet{li2025treepo} propose to use a process search to branch out the trajectories and collect data.


\paragraph{Addressing signal loss in GRPO.} A central challenge in applying GRPO is the ``signal loss'' problem, where groups of responses with uniform rewards yield zero gradients. Prior work has identified this issue and proposed several solutions. The most direct approach is passively filtering out these prompts \citep{yu2025dapo, xiong2025minimalist}. Another line of works proposes to modify the advantage  computation and avoid zero gradient. To avoid discarding samples, \citet{nan2025ngrpo} propose augmenting the reward group with a constant (the maximum possible reward), ensuring the variance is not zero by introducing a bias. \citet{le2025no} propose to assign advantages based on entropy information for prompts with uniform rewards. Finally, some works propose to mitigate this issue by selectively choose the prompt batch. For instance, \citet{qu2025can} employ a Bayesian framework to predict a prompt's pass rate and selectively sample informative prompts during online training. Our work differs by tackling the problem at the collection stage itself, ensuring a sufficient signal is gathered adaptively rather than correcting for its absence afterward. Similarly, \citet{shi2025efficient} propose to use an adaptive  curriculum learning to select prompts with suitable difficulty during the online RL training. \citet{zhang2025speed} also develop a curriculum learning methods to select training prompts of intermediate difficulty in a two-stage manner. \citet{zheng2025act} use a dictionary-based approach to record historical reward from the last epoch and skip uninformative prompts. 

\paragraph{Learning objective in RLVR.} A key contribution of this work is the systematic formulation of non-linear RL objectives and the weighted gradient estimators that arise from them. Our theoretical framework builds directly on the foundational insights of \citet{yao2025optimizing}, which is one of the earliest works to recognize that RLVR can be viewed through a prompt-dependent weighting scheme from a theoretical perspective. In particular, \citet{yao2025optimizing} interpret rejection sampling fine-tuning (RAFT) \citep{dong2023raft} under an Expectation-Maximization (EM) view \citep{singh2023beyond, zhong2025brite} and make the important observation that the true expected gradient in EM-RAFT is weighted by $1/p_\theta(x)$. They further characterize the variance structure under this weighted gradient and show that the optimal sampling budget should scale with $1/\sqrt{p_\theta(x)}$. Motivated by their results, our early attempts tried to extend their optimal variance-reduction estimator to the standard {Reinforce} algorithm, but this turned out to be infeasible. This led us to analyze the log-objective for {Reinforce}, which naturally recovers a variance structure similar to EM-RAFT. Our framework also captures and generalizes the result of \citet{yao2025optimizing} (see Appendix~\ref{appendix:vr} for details). 

From an algorithmic perspective, \citet{yao2025optimizing} propose an improved RAFT variant with variance-reduced gradient estimates. However, their method follows a two-stage ``explore-then-exploit'' procedure: a small portion of the budget is used to estimate pass rates (often offline), and the remaining budget is allocated based on these estimates. This design has a key limitation—when the initial budget is small, the pass-rate estimate has very high variance, especially for difficult prompts (low $p_i$), where the initial sample almost always contains zero positive responses. In contrast, we study the more general {Reinforce} algorithm, clarify the connection between explicit weighting and implicit weighted sampling, and provide two practical realizations—an estimation-based method and a model-free sequential sampling method—that mitigate this issue.

We also note a very recent theoretical note by \citet{davis2025objective}, which appeared concurrently with our second pre-print version. Their work also observes that non-linear objectives can be linked to heuristic weighting schemes used in algorithms such as GRPO. In comparison, we not only establish the theoretical link but also explicate the duality between optimizing this objective via explicit weights versus adaptive sampling. Furthermore, we go beyond theory to translate these insights into a practical, efficient algorithm, \textsc{Reinforce-Ada}, and rigorously validate its effectiveness through extensive experiments.

\paragraph{GRPO variant designs.}
Our work is orthogonal to another line of research focusing on modifying the policy gradient algorithm itself, such as innovations in advantage estimation and clipping mechanisms \citep{hu2025reinforce++, zhu2025surprising, zheng2025group, huang2025mapo, chu2025gpg}. While these methods refine the core update rule, we focus on data generation and construction pipeline. Our adaptive sampling framework is complementary to these algorithmic improvements and is possible to be combined with them.

\section{Discussion and End Note}
In this work, we presented a general framework for optimizing non-linear RL objectives, identifying adaptive sampling as a robust strategy to address the signal loss problem inherent in standard Reinforce with group sampling algorithm. Guided by this framework, we introduced \textsc{Reinforce-Ada}, a family of algorithms that dynamically allocate inference budgets based on prompt difficulty. By prioritizing prompts that require more exploration, our method efficiently constructs informative training groups, recovering valid learning signals that are typically lost in uniform sampling. Designed as a \textit{lightweight, drop-in replacement} for the standard RL training loop, \textsc{Reinforce-Ada} delivers \textit{consistent and robust improvements} across diverse foundation models, accelerating convergence and boosting final performance. These benefits come with only moderate computational overhead, offering a scalable and theoretically grounded alternative to the brute-force approach of uniformly large group sizes.

We view this work as part of the broader recent effort on \textbf{data curation for online reinforcement learning}. Recent studies have explored \textit{macro, prompt-level} strategies, such as curriculum learning \citep{zhao2024automatic, shi2025efficient, zhang2025speed}, to shape the distribution of training data during online learning. In contrast, our contribution operates at the \textit{response-sampling level}, focusing on how to construct effective learning signals within each prompt. However, as discussed in Section~\ref{sec:dynamic_and_ablation}, the relative difficulty of the prompt set evolves alongside model training, and this interplay critically affects both learning dynamics and final performance of our method. Moreover, while our experiments are restricted to the math domain due to resource constraints, real-world post-training systems require data curation as a \textit{holistic challenge} spanning the entire pipeline. We hope that the adaptive sampling framework can serve as an effective building block in this broader ecosystem when combined with complementary approaches from the literature.

\section*{Acknowledgment}
The authors would like to thank Ziniu Li for the insightful discussions.

\bibliography{main}
\bibliographystyle{tmlr}

\newpage
\appendix

\section{Authorship and Credit Attribution}
\label{appendix:credits}
All authors provided valuable contributions to this project, each bringing unique expertise and insights that were crucial for its success. 

\textbf{WX} proposed the project idea, initiated and organized the project, and developed the core sequential sampling algorithm; developed the framework of non-linear RL objective; implemented the initial codebase, processed the data, and ran proof-of-concept experiments to validate its effectiveness; drafted the initial version of the paper, with subsequent contributions and revisions from co-authors.

\textbf{CY} proposed the idea of global normalization; jointly developed the core sequential sampling algorithms, including the codes for Reinforce-Ada-Seq-pos, Reinforce-Ada-Seq-balance, global normalization, and verification; ran the experiments for Qwen2.5-Math-1.5B and Qwen2.5-Math-7B; contributed to the paper writing.

\textbf{BL} processed the data; ran the experiments for Qwen2.5-Math-7B, Llama, Qwen3-4B; provided the released version of code (including the Tinker version); and contributed to the paper proofreading.

\textbf{HD} initiated, coordinated, and drove the project;
provided insights about the algorithm and project design;
implemented the production-ready \textsc{Reinforce-Ada}; developed and refactored a scalable and configurable codebase; conducted key experiments for Qwen2.5-Math-1.5B; made substantial contributions to the manuscript writing.

\textbf{XX}, \textbf{CM}, \textbf{JB}, \textbf{NJ}, \textbf{TZ} are senior authors, supported and advised the work, provided resources, and suggested experiments and improvements to the writing.

\section{Variance-Reduction Gradient Estimator under the Log-objective}
\label{appendix:vr}

In this section, we connect our general framework to the variance-reduction principle explored in \citet{yao2025optimizing}. We generalize their findings from the specific case of Rejection Sampling Fine-Tuning (RFT) to the general Reinforce algorithm with a baseline.

We recall that \citet{yao2025optimizing} demonstrated that RFT can be viewed as maximizing the log-objective $J_{\log}(\theta) = \mathbb{E}[\log p_\theta(x)]$. Here, we analyze the gradient estimator for this objective under the general Reinforce algorithm with baseline:
\begin{equation}
    g_{\log}(x) = \frac{1}{p(x)} \cdot \E_{a \sim \pi_\theta(\cdot|x)} \Big[ \nabla_\theta \log \pi_\theta(a|x) \cdot (r(x, a) - p(x) ) \Big].
\end{equation}

\subsection{Optimal Budget Allocation}

Our goal is to estimate the total gradient for a batch $g_{batch} = \sum_i g_{\log}(x_i)$ with minimum variance, given a total budget $N$. 
Suppose we have a total of $N$ samples to distribute across $B$ prompts $x_1,\dots,x_B$. Using $n_i$ samples for prompt $x_i$, the batch gradient estimator is
\begin{equation} \label{eqn:batch_vr}
    \hat{g}_{batch} = \frac{1}{B} \sum_{i=1}^B \frac{1}{n_i p_i} \sum_{j=1}^{n_i} \nabla_\theta \log \pi(a_{ij}|x_i)\,(r_{ij}-p_i),
\end{equation}
where $p_i=p_\theta(x_i)$.

\begin{figure}[htp]
    \centering
        \includegraphics[width=0.7\textwidth]{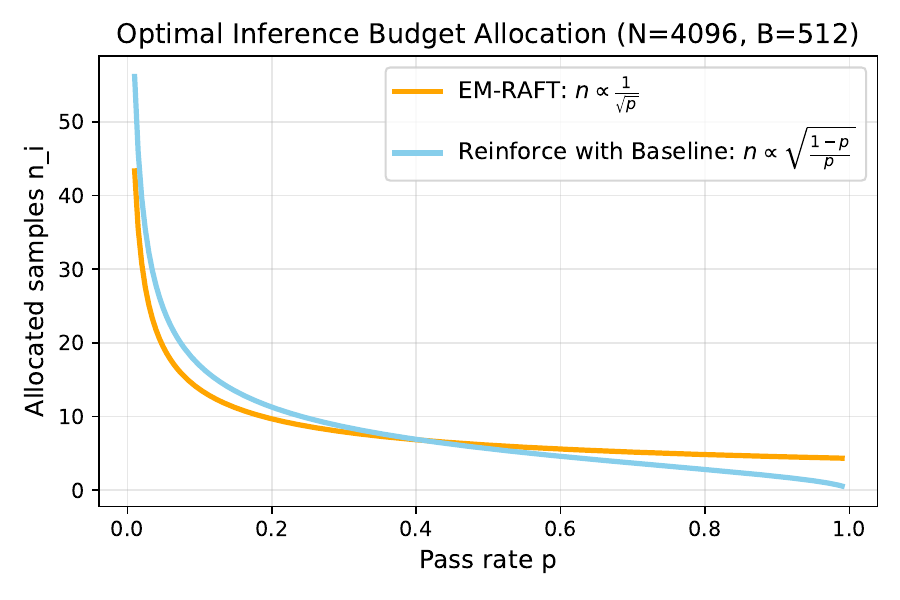}  
        \caption{
                The theoretically optimal inference budget allocation of RAFT and Reinforce with baseline algorithms under log-objective.
    }
        \label{fig:est_cost} 
\end{figure}

\textbf{Crucially, note the distinction from Section \ref{sec:theory}:} In this estimator, the optimization of the log-objective is captured entirely by the explicit factor $1/p_i$. Since the inner sum is normalized by $n_i$, the allocation $n_i$ does not affect the optimization objective (it is an unbiased estimator for any $n_i \ge 1$). Therefore, the role of $n_i$ here is purely to minimize the variance of the estimator. We seek the allocation $(n_1,\dots,n_B)$ minimizing the total gradient variance:
\begin{align} \label{eqn:opt_problem}
\min_{n_1,\dots,n_B} &\quad V = \sum_{i=1}^B \frac{\sigma_g^2(x_i)}{n_i p_i^2}, \\
    \text{subject to} \quad & \sum_{i=1}^B n_i = N \quad \text{and} \quad n_i \ge 1. \nonumber
\end{align}
Here $\sigma_g^2(x_i)$ is the variance of the baselined Reinforce gradient $\nabla_\theta \log \pi(a_{ij}|x_i)(r_{ij}-p_i)$. Assuming gradient norms are similar across prompts (see the derivation of Proposition 1 in \citet{yao2025optimizing} for details), we have 
\[
\sigma_g^2(x) \propto
\begin{cases}
p(x) & \text{(RFT without baseline)},\\
p(x) \cdot (1-p(x)) & \text{(Reinforce with baseline)}.
\end{cases}
\]
Solving \eqref{eqn:opt_problem} via Lagrange multipliers gives the optimal allocation rule:
\begin{equation}\label{eqn:alloc_rule}
    n_i^\star \propto \frac{\sigma_g(x_i)}{p_i},
\end{equation}
which yields
\[
\text{RFT: } n_i^\star \propto \sqrt{\tfrac{1}{p_i}}, \qquad
\text{Reinforce: } n_i^\star \propto \sqrt{\tfrac{1-p_i}{p_i}}.
\]
This shows that incorporating a baseline reduces gradient variance and thus changes the optimal sampling strategy. Combining these findings, we can formulate a reasonable variance-reduction algorithm.

\paragraph{Reinforce with baseline.}  
At each iteration $t$:
\begin{enumerate}
    \item Allocate inference budgets as $n_i \propto N \cdot \sqrt{\tfrac{1-p_{\theta_t}(x_i)}{p_{\theta_t}(x_i)}}$.
    \item For each prompt $x_i$, compute the advantage:
    \[
        A^t(x_i,a_{ij}) = \frac{1}{n_i p_i} \cdot (r_{ij}-p_{\theta_t}(x_i)) \propto \frac{1}{\sqrt{p_{\theta_t}(x_i)\cdot (1-p_{\theta_t}(x_i))}}\cdot (r_{ij}-p_{\theta_t}(x_i)).
    \]
    \item Update the policy using the weighted gradient.
\end{enumerate}

\begin{remark}[Recovering the GRPO Advantage]
The derived advantage term $\frac{r - p}{\sqrt{p(1-p)}}$ is exactly equivalent to normalizing the reward by its standard deviation, since $\text{Std}[r] = \sqrt{p(1-p)}$ for binary rewards. This result provides an interesting explanation for the heuristic used in algorithms like GRPO \citep{shao2024deepseekmath}, which normalize advantages by the group standard deviation. 
\end{remark}

\begin{remark}[The Necessity of the Log-Objective]
It is important to note that the log-objective factor ($1/p$) in \eqref{eqn:batch_vr} is essential for this result. If we were optimizing the standard objective $J=\mathbb{E}[p]$ (where the weight is 1), the variance term would be $\frac{\sigma_g^2(x_i)}{n_i}$ instead of $\frac{\sigma_g^2(x_i)}{n_i p_i^2}$. Minimizing that variance would lead to allocating more budget to prompts with maximum variance (i.e., $p \approx 0.5$) rather than the difficult prompts ($p \to 0$). Thus, the log-objective is a prerequisite for focusing the variance-reduction effort on difficult problems.
\end{remark}

\section{Ablation Studies}

\subsection{Adaptive Sampling Solves More Difficult Prompts}
\label{sec:diff_level}

\begin{table*}[htp]
    \centering
    \begin{adjustbox}{max width=\textwidth}
    \begin{tabular}{clccccc} 
    \toprule
        Model & Algorithm & \textbf{Math500} & \textbf{Minerva Math} & \textbf{Olympiad Bench} & \textbf{AIME-like} & \textbf{Weighted Average} \\
        \midrule
        \multirow{3}{*}{\textit{Qwen2.5-Math-1.5B}} 
        & GRPO-n4 & 74.2 & 34.4 & 38.4 & 16.2 & 45.3\\
        & \textsc{Reinforce-Ada-Seq-pos} & 75.8 & 35.7 & 38.6 & 16.5 & 46.1 (+0.8)\\
        & \textsc{Reinforce-Ada-Seq-balance} & 77.4 & 36.5 & 40.5 & 17.5 & \textbf{47.6 (+2.3)}\\
            \midrule
        \multirow{3}{*}{\textit{Qwen2.5-Math-1.5B (hard)}} 
        & GRPO & 71.0 & 31.8 & 34.3 & 13.8 & 41.9\\
        & \textsc{Reinforce-Ada-Seq-pos} & 73.9 & 33.1 & 36.4 & 16.4 & 44.6 (+2.7)\\
        & \textsc{Reinforce-Ada-Seq-balance} & 74.7 & 33.7 & 38.7 & 17.6 & \textbf{45.5 (+3.6)} \\
        \midrule
        \multirow{3}{*}{\textit{Qwen2.5-Math-7B}} 
        & GRPO & 82.2 & 44.7 & 45.6 & 23.2 & 53.3\\
        & \textsc{Reinforce-Ada-Seq-pos} & 82.7 & 45.1 & 46.7 & 23.7 & 54.2 (+0.9)\\
        & \textsc{Reinforce-Ada-Seq-balance} & 84.0 & 45.2 & 47.1 & 23.7  & \textbf{54.6 (+1.3)}\\
           \midrule
        \multirow{3}{*}{\textit{Qwen2.5-Math-7B (hard)}} 
        & GRPO & 80.7 & 42.8 & 42.9 & 21.8 & 51.3\\
        & \textsc{Reinforce-Ada-Seq-pos} & 82.4 & 43.1 & 45.0 & 22.2 & 52.8 (+1.5) \\
        & \textsc{Reinforce-Ada-Seq-balance} & 83.1 & 43.4 & 46.4 & 24.9 & \textbf{53.9 (+2.6)} 
        \\
    \bottomrule
    \end{tabular}
    \end{adjustbox}
        \caption{Ablation study on the impact of prompt difficulty. The gains become more pronounced on the more challenging training set, where our methods allocate more inference budget to difficult prompts to recover a valid learning signal, while GRPO samples all prompts uniformly.}    \label{tab:abl_difficulty}
\end{table*}

\begin{figure}[htp]
    \centering
    \includegraphics[width=0.49\textwidth]{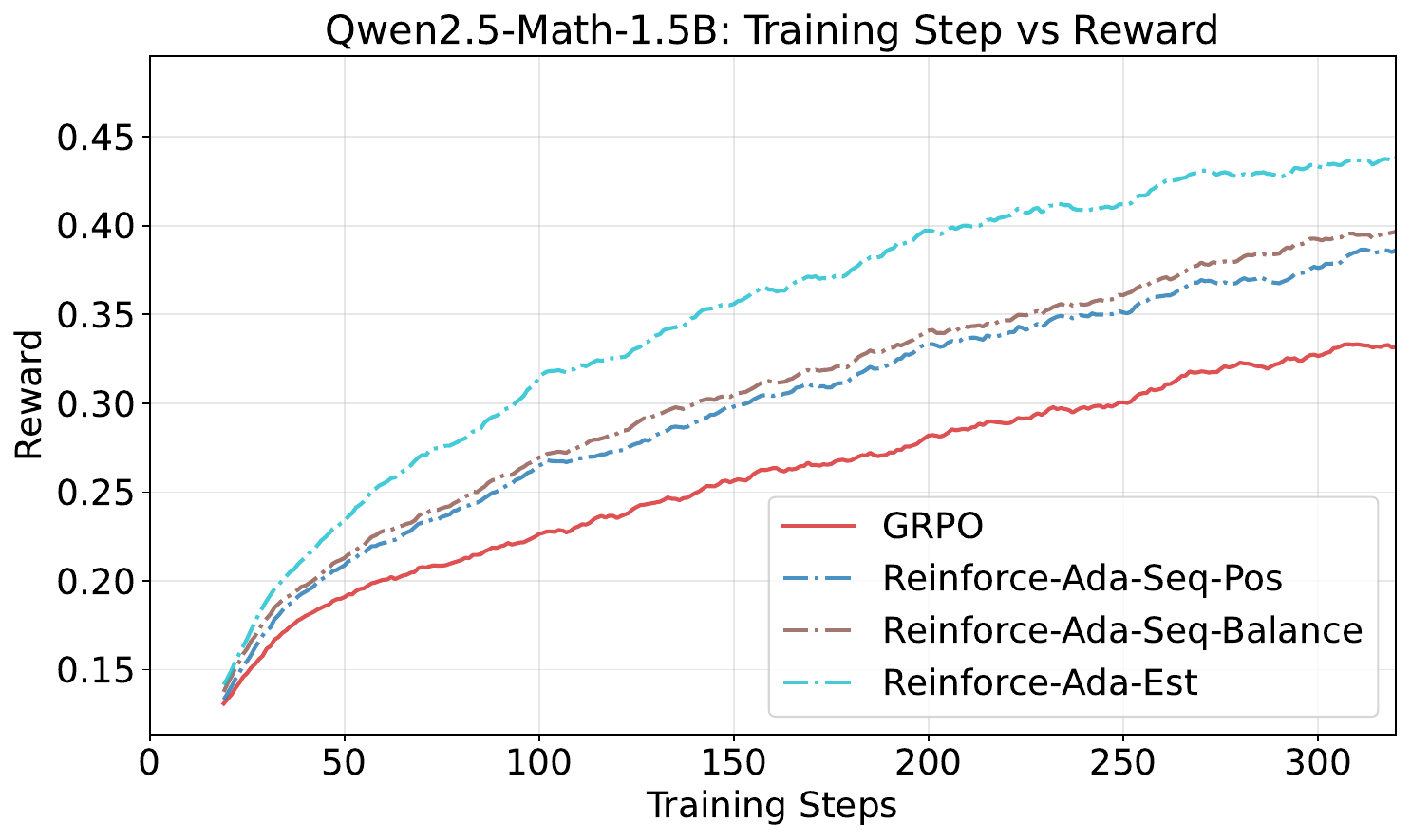}  
    \includegraphics[width=0.49\textwidth]{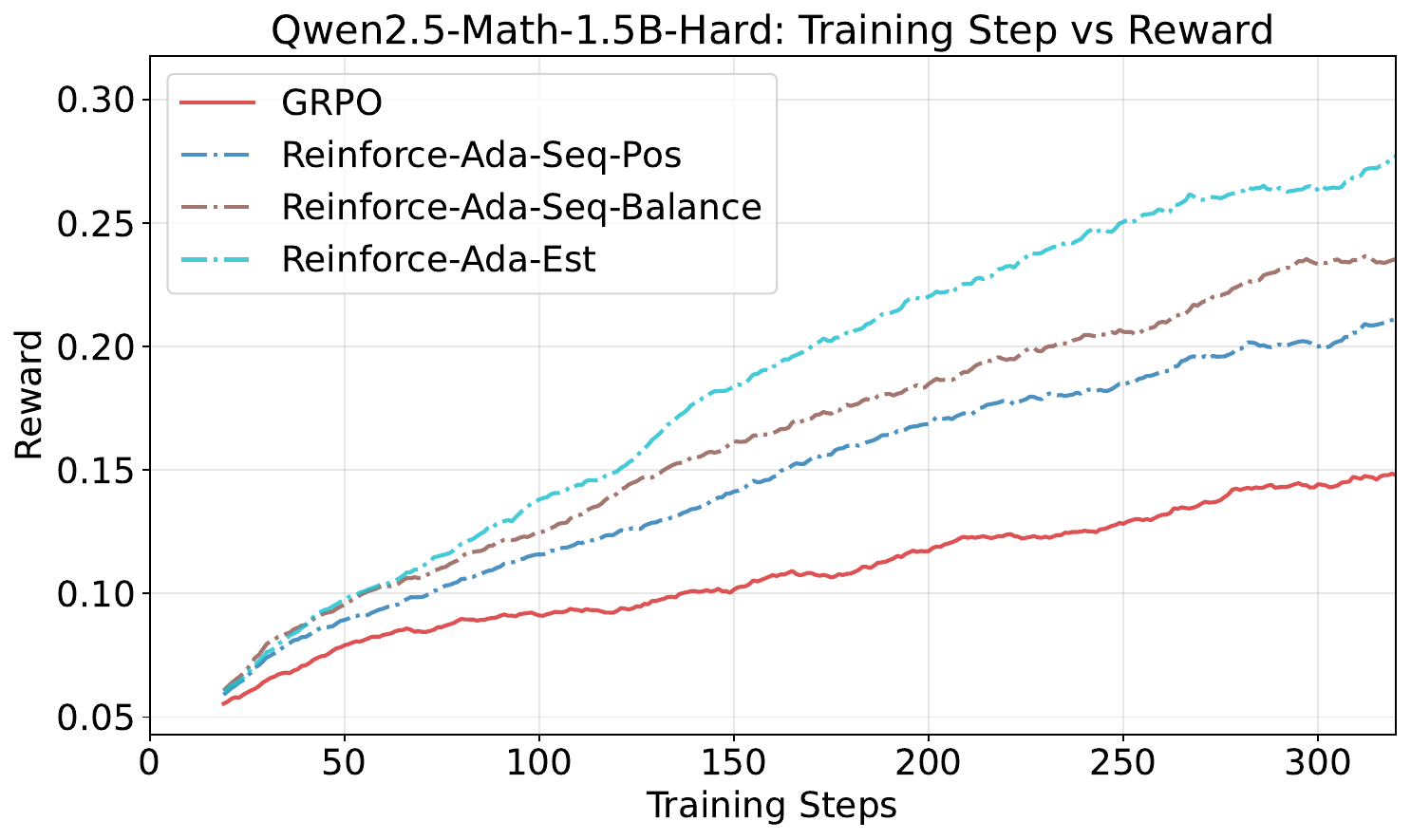}
    \caption{Ablation study on the prompt set difficulty. Left: prompt set with moderate difficulty. Right: challenging prompt set. The benefit of adaptive sampling is more obvious with challenging prompt set.}
        \label{fig:difficulty_ablation} 
\end{figure}

\paragraph{The impact of prompt set difficulty} 
After $\sim$200 steps (Fig.~\ref{fig:sampling_dynamic}, right, second row), most prompts become easy—two rounds already satisfy the positive quota—so additional sampling brings diminishing returns. To stress-test our method, we construct a hard subset that keeps only prompts with 1-2 correct out of 16 initial samples. As shown in Fig.~\ref{fig:difficulty_ablation}, adaptive sampling yields a much larger margin over GRPO on this challenging set, and the gap widens late in training. This is also reflected on the final model performance reported in Table~\ref{tab:abl_difficulty}.

Intuitively, harder prompts reduce all-correct saturation and keep uncertainty high; \textsc{Reinforce-Ada} continues to mine informative negatives and avoids premature deactivation, sustaining exploration and gradient signal.

\begin{table}[t]
\centering
\small
\setlength{\tabcolsep}{4.5pt}
\begin{tabular}{lccccc}
\toprule
\textbf{Method} & 
\textbf{Extremely Hard} & 
\textbf{Hard} & 
\textbf{Medium} & 
\textbf{Easy} & 
\textbf{Overall} \\
\midrule
Base Model &
0.0000 & 0.0889 & 0.2950 & 0.6151 & 0.1139 \\

GRPO &
0.3414 {\scriptsize(+34.14)} &
0.4639 {\scriptsize(+37.51)} &
0.6497 {\scriptsize(+35.46)} &
0.6865 {\scriptsize(+7.14)} &
0.4749 {\scriptsize(+36.11)} \\

DAPO &
0.3487 {\scriptsize(+34.87)} &
0.4648 {\scriptsize(+37.60)} &
0.6521 {\scriptsize(+35.71)} &
0.5580 {\scriptsize(-5.70)} &
0.4757 {\scriptsize(+36.19)} \\

\textbf{Reinforce-Ada} &
\textbf{0.3674} {\scriptsize(+36.74)} &
\textbf{0.4825} {\scriptsize(+39.37)} &
\textbf{0.6579} {\scriptsize(+36.29)} &
\textbf{0.7153} {\scriptsize(+10.02)} &
\textbf{0.4933} {\scriptsize(+37.94)} \\
\bottomrule
\end{tabular}
\caption{Accuracy and absolute improvements over the base model across difficulty levels. \textsc{Seq-Balance} achieves consistent improvements across all difficulty categories.}
\label{tab:difficulty-swapped-improvements}
\end{table}

To better understand how adaptive allocation reshapes the learning dynamics, we randomly sample a subset of 5000 prompts from the hard training set. We partition the 5000-prompt subset into four difficulty levels using the base model’s pass@16: (i) Extremely Hard (0), (ii) Hard (0–0.2], (iii) Medium (0.2–0.5], and (iv) Easy (0.5–1.0]. The dataset is dominated by difficult items where 984 (19.7\%) are extremely hard and 3,082 (61.7\%) are hard. Meanwhile, medium and easy prompts account for 870 (17.4\%) and 63 (1.3\%), respectively. As expected, the base model’s accuracy rises monotonically with difficulty, achieving 0.00\%, 8.89\%, 29.50\%, and 61.51\% on the four groups.

Table \ref{tab:difficulty-swapped-improvements} summarizes different models' accuracy across four difficulty levels. All three RL methods substantially improve over the base model, but \textsc{Reinforce-Ada-Seq-Balance} achieves the highest accuracy in every category. The gains are especially pronounced on the Extremely Hard and Hard prompts—where the base model succeeds rarely or not at all—improving accuracy by +36.74 and +39.37 points, respectively. \textsc{Reinforce-Ada-Seq-Balance} also attains the best performance on Easy prompts (71.53\%), whereas DAPO even hurts the performance in this regime. A likely reason is that DAPO tends to discard high-pass-rate prompts during training, while our balanced adaptive sampling variant allocates sufficient inference budget to preserve learning signals even for easy items.

\subsection{Variants of \textsc{Reinforce-Ada-Est}}
\label{appendix:abl_hybrid}
We also compare the effect of sampling and reweighting in \textsc{Reinforce-Ada-Est}. The training reward averaged at the prompt level is similar for both the hybrid strategy ($w_{\mathrm{sample}} = 1/\sqrt{p}, w_{\mathrm{grad}} = 1/\sqrt{p}$) and the implicit weighting via sampling  ($w_{\mathrm{sample}} = 1/p, w_{\mathrm{grad}} = 1$). However, the batch-level reward under implicit weighting eventually decreases. This indicates that the method is overly aggressive in sampling negative samples, leading to excessive oversampling of difficult prompts. When the task distribution contains many hard examples, this issue becomes more pronounced, since simpler samples receive too few draws.

In fact, the number of samples drawn per prompt $N_x$, the total sampling budget $N_\mathrm{total}$, and the reward $r_x$ satisfy
$$
(r_\mathrm{batch} - r_\mathrm{prompt}) N_\mathrm{total} = \mathbf{Cov}(N_x,r_x),
$$
A large gap between prompt level reward and batch level reward implies a strong negative covariance between $N_x$ and $r_x$, meaning low reward prompts dominate the sampling budget. Under a finite sampling constraint, it is therefore important to ensure broad prompt coverage and use a more moderate sampling strategy to avoid such imbalance and extreme behavior.
\begin{figure}[htp]
    \centering
    \includegraphics[width=0.48\textwidth]{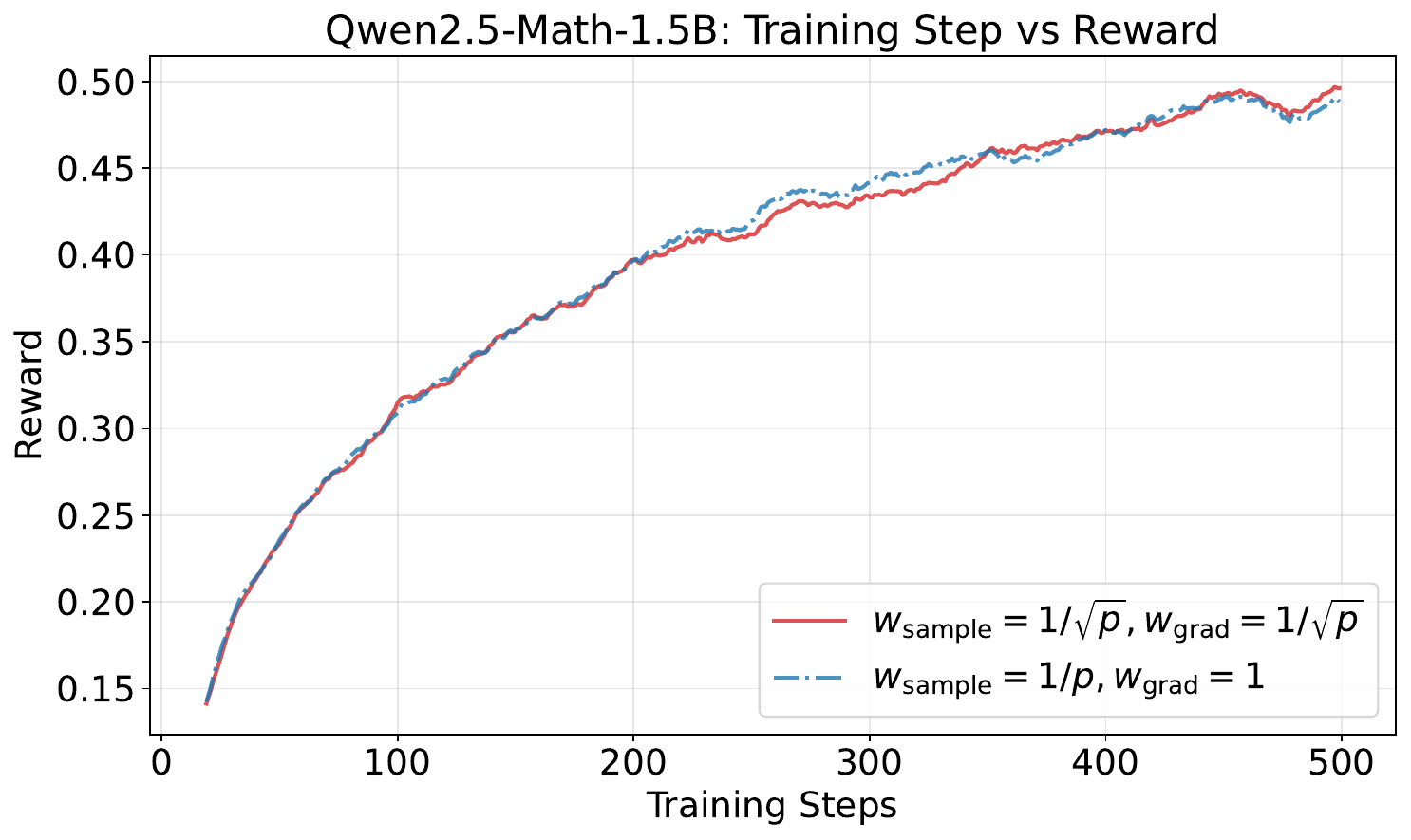} 
    \includegraphics[width=0.48\textwidth]{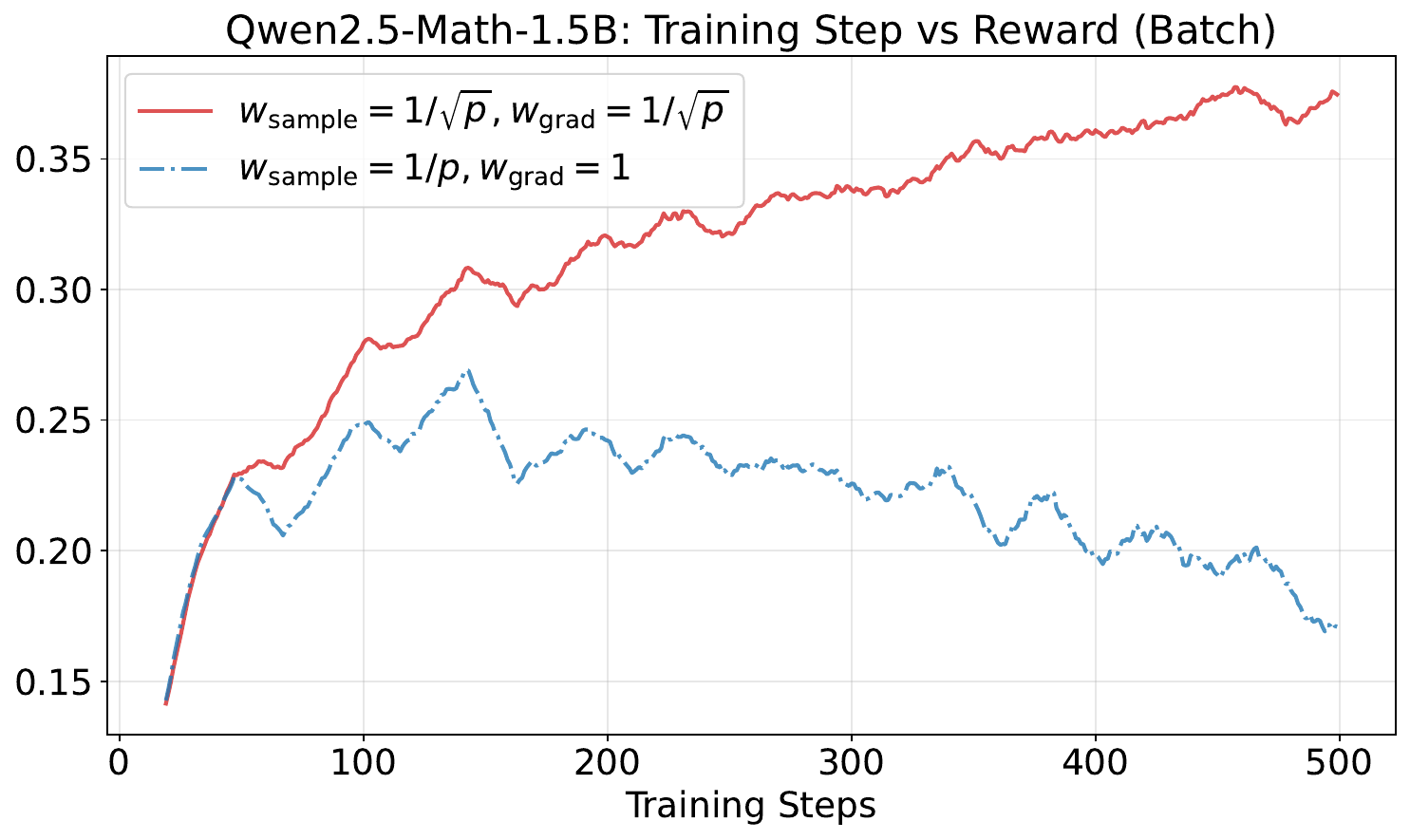}  
\caption{Comparison with different sampling/reweighting strategy.
}
        \label{fig:est-variants} 
\end{figure}
\section{Simulation of Sampling Cost of {Reinforce-Ada-Seq}}

\begin{figure}[htp]
    \centering
        \includegraphics[width=0.32\textwidth]{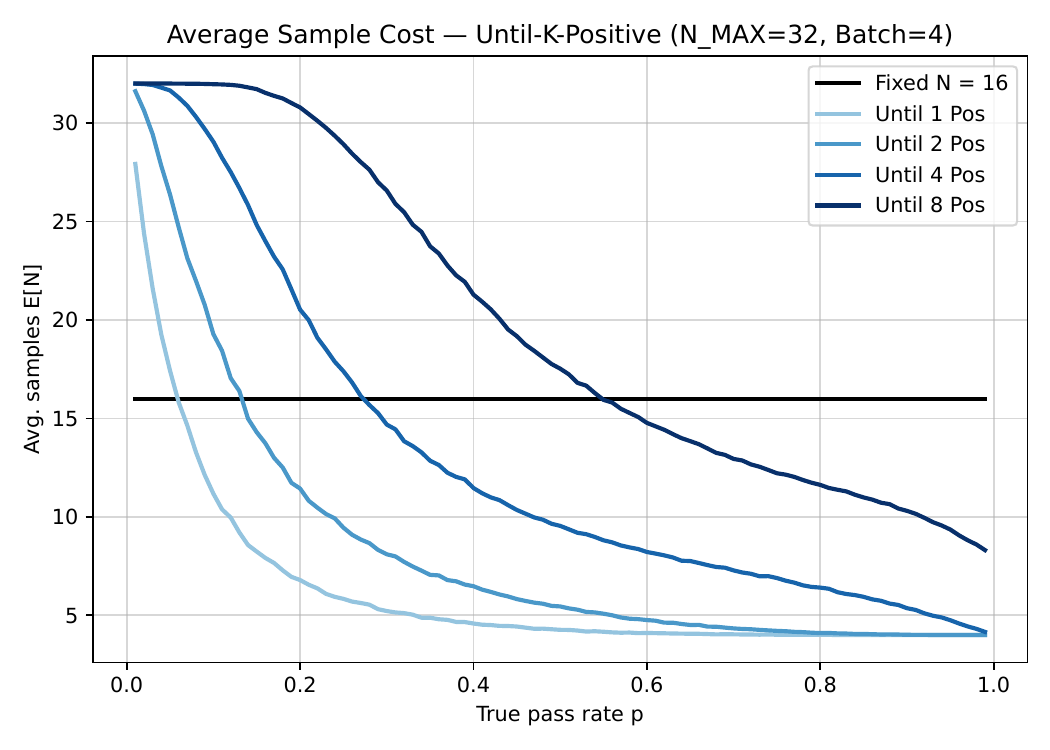}  
    \includegraphics[width=0.32\textwidth]{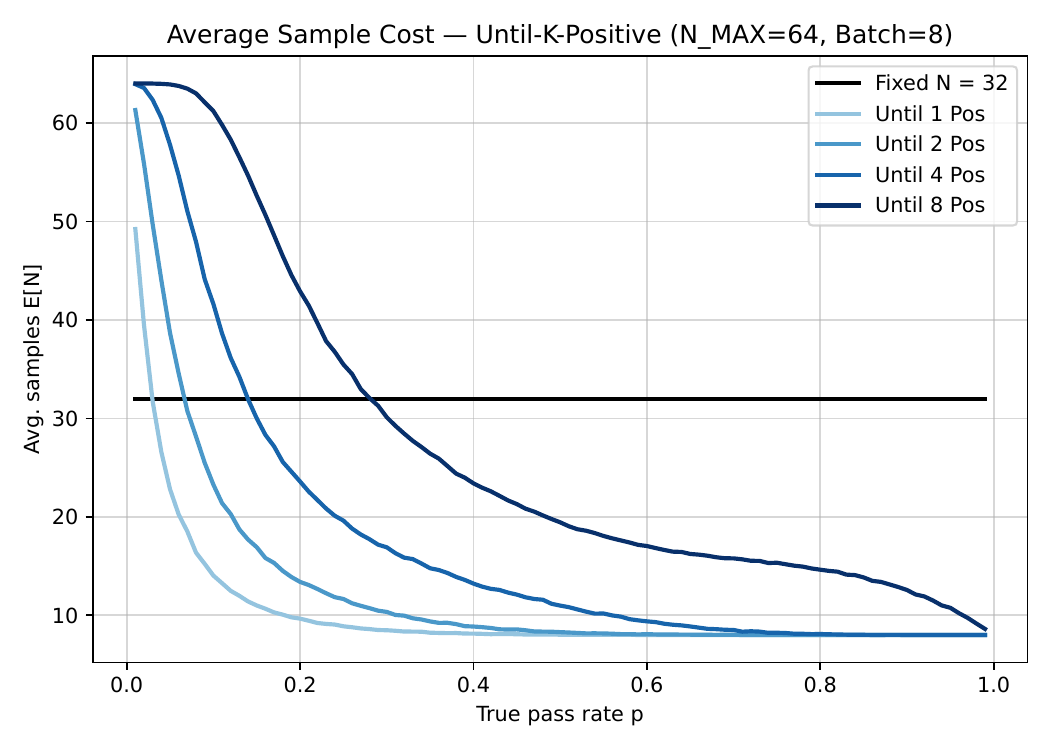}
    \includegraphics[width=0.32\textwidth]{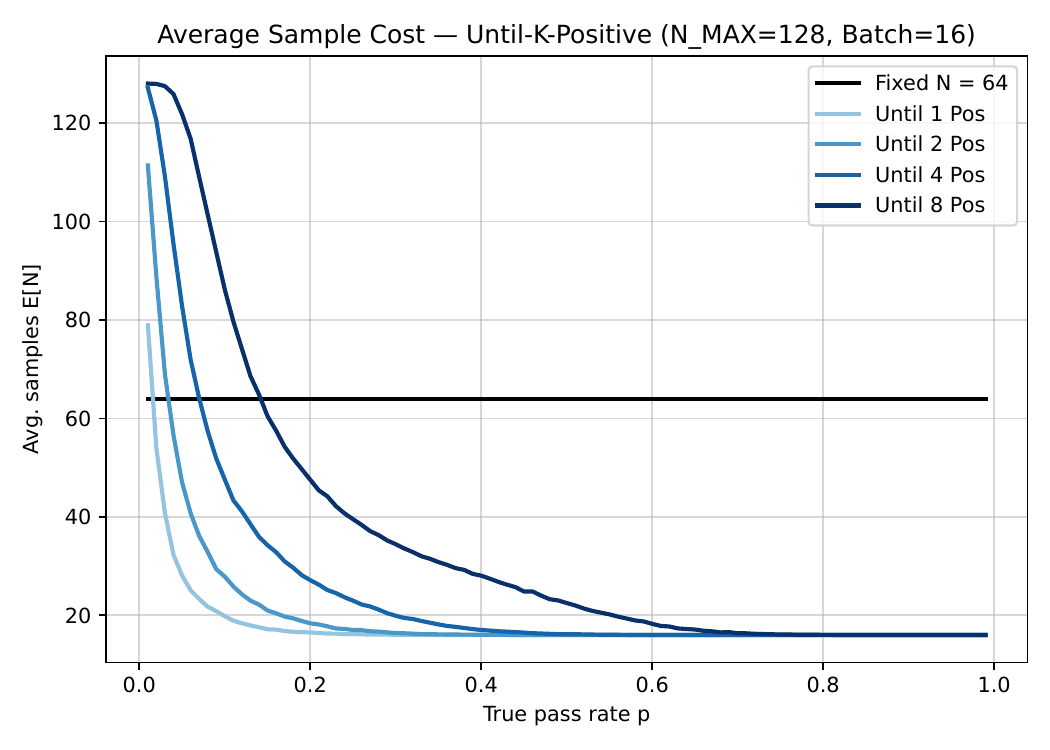}
    \includegraphics[width=0.32\textwidth]{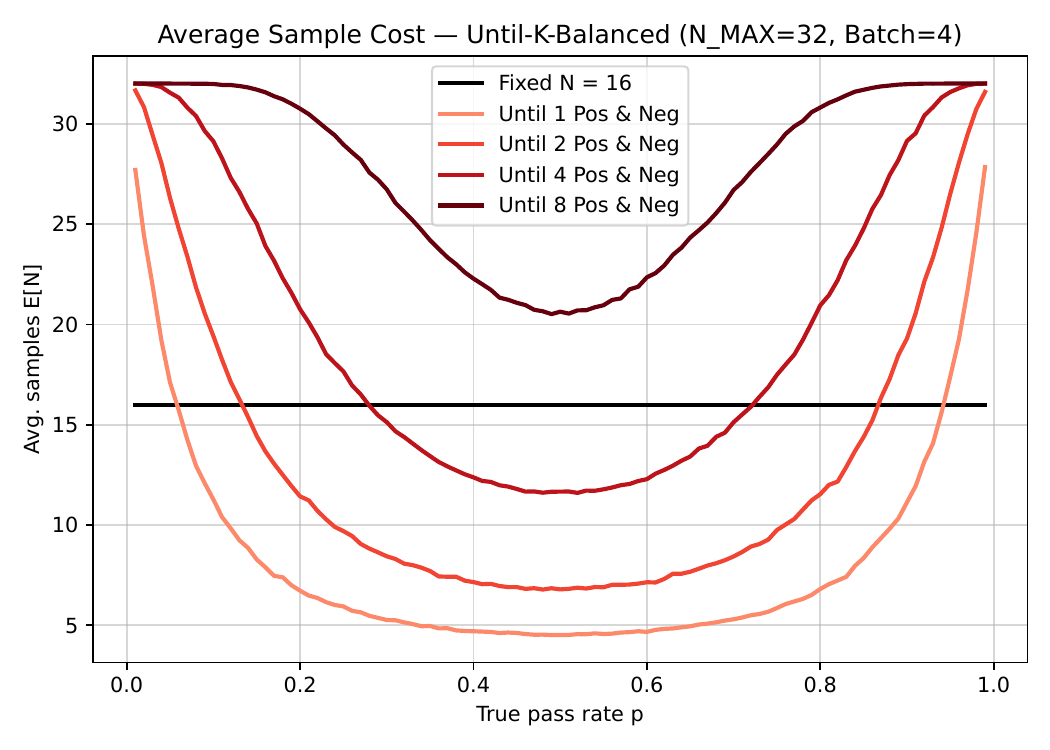}  
    \includegraphics[width=0.32\textwidth]{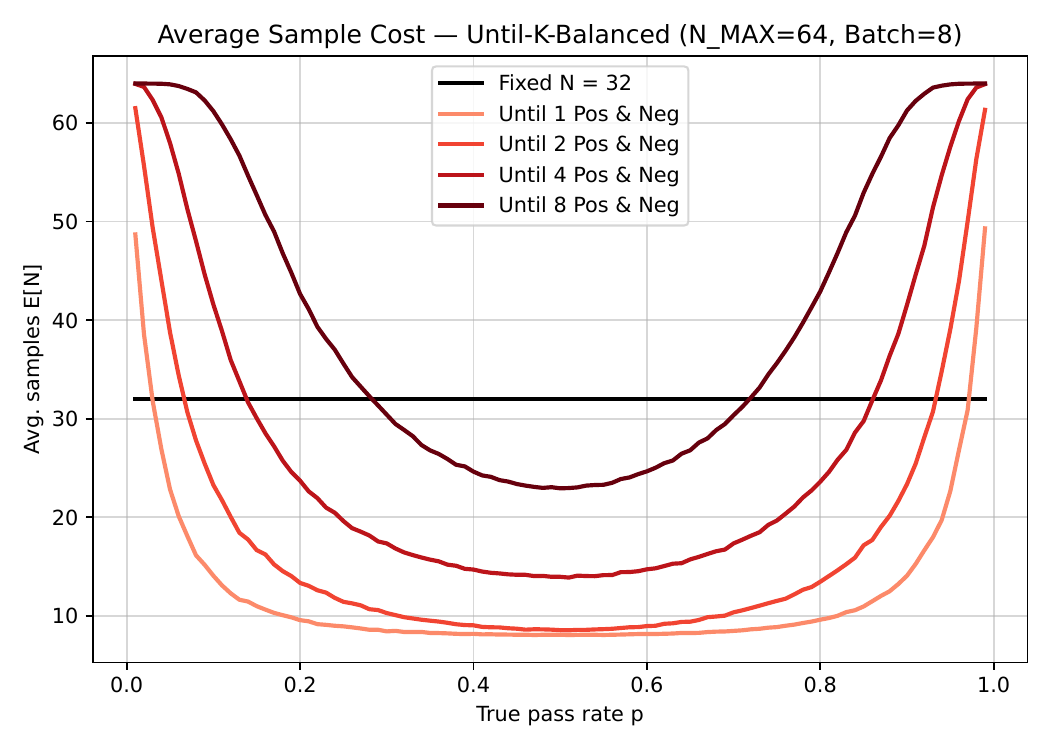}
    \includegraphics[width=0.32\textwidth]{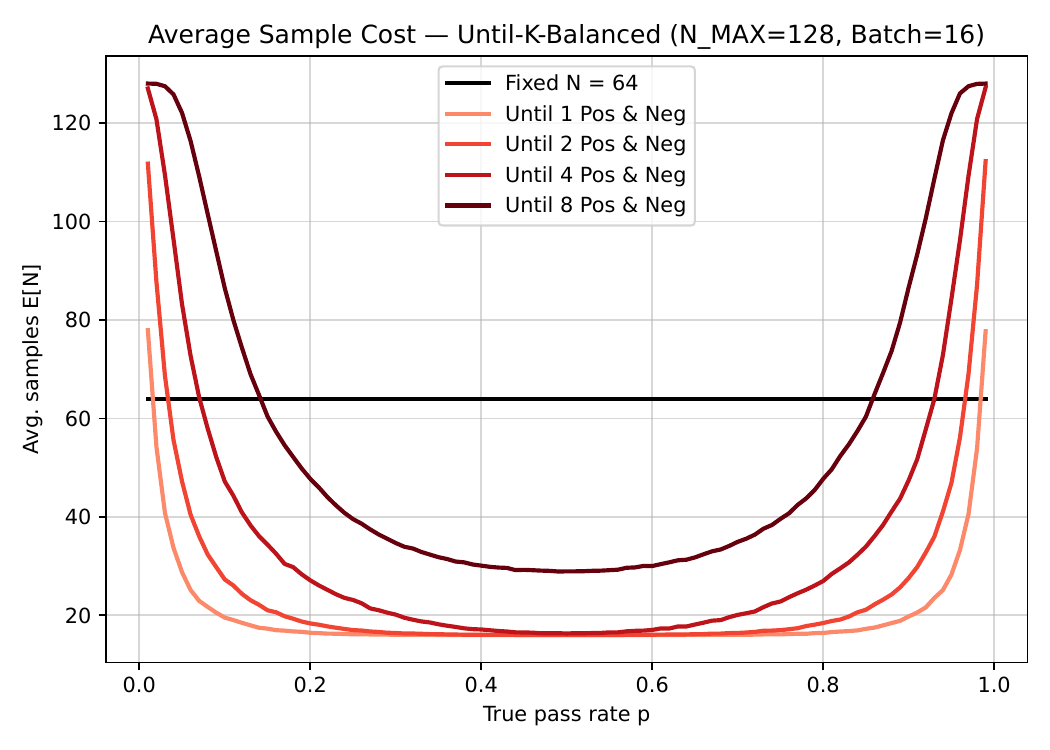}
        \caption{
         Estimated sampling cost under sequential stopping rules.
        The first row shows the expected number of responses required until observing $K$ positive samples; 
        the second row shows the cost until collecting $K$ positive and $K$ negative samples. 
        Each column corresponds to a different maximum exploration budget $N_{\max}\in\{32,64,128\}$. 
    }
        \label{fig:est_cost2} 
\end{figure}

To better understand the sampling cost induced by our sequential stopping rules, we conduct a set of numerical simulations illustrated in Figure~\ref{fig:est_cost2}. We evaluate two variants of our method: \textsc{Reinforce-Ada-Seq-pos}, which continues sampling until $K$ positive responses are observed, and \textsc{Reinforce-Ada-Seq-balance}, which stops once both $K$ positive and $K$ negative responses have been collected. 

For each strategy, we consider three maximum exploration budgets $N_{\max} \in \{32, 64, 128\}$ and use a batch size of $k_{\text{batch}} = N_{\max}/8$. We then vary the stopping requirement~$K$ and estimate the expected number of samples required as a function of the true success probability $p \in (0, 1)$. Because each response is a Bernoulli trial, the total number of required samples can be simulated accurately and efficiently.

\end{document}